\documentclass[10pt,twocolumn,letterpaper]{article}

\usepackage{cvpr}
\usepackage{times}
\usepackage{epsfig}
\usepackage{graphicx}
\usepackage{amsmath}
\usepackage{amssymb}
\usepackage{enumitem}
\usepackage{amsmath}
\usepackage{balance}
\usepackage{stfloats}
\DeclareMathOperator*{\argmax}{argmax} % thin space, limits underneath in displays
\DeclareMathOperator{\IoU}{IoU}
\usepackage{ml-acro}
\usepackage{mathbbol}
\usepackage{subcaption}
\usepackage{dirtytalk}
\usepackage{ftnxtra}
\renewcommand{\thesubfigure}{\alph{subfigure}}
\newcommand{\mycaption}[1]% #1 = caption
{\refstepcounter{subfigure}\textbf{(\thesubfigure) }{\ignorespaces #1}}

\usepackage{booktabs}
\usepackage{textcomp}
 \cvprfinalcopy

\usepackage[breaklinks=true,bookmarks=false]{hyperref}
\makeatletter
\renewcommand\subsubsection{\@startsection{subsubsection}{3}{\z@}%
                                     {-3.25ex\@plus -1ex \@minus .2ex}%
                                     {-1em}%
                                     {\normalfont\normalsize\bfseries}}
\makeatother
\cvprfinalcopy % *** Uncomment this line for the final submission

\def\cvprPaperID{1628} % *** Enter the CVPR Paper ID here

\ifcvprfinal\fi
\begin{document}

%%%%%%%%% TITLE
\title{Real-Time Panoptic Segmentation from Dense Detections}
\author{Rui Hou$^{*,1, 2}$ \quad
          Jie Li$^{*,1}$\quad
   Arjun Bhargava$^{1}$\quad
   Allan Raventos$^{1}$ \quad
   Vitor Guizilini$^{1}$ \quad
   Chao Fang$^1$ \\
Jerome Lynch$^{2}$ \quad
   Adrien Gaidon$^1$ 
\and
$^1$Toyota Research Institute\quad
$^{2}$University of Michigan, Ann Arbor\\
{\tt\small $^1$\{firstname.lastname\}@tri.global}\quad
{\tt\small $^{2}$\{rayhou, jerlynch\}@umich.edu}\quad
% Institution2\\
% First line of institution2 address\\
% {\tt\small secondauthor@i2.org}
}

\maketitle
\thispagestyle{empty}

%%%%%%%%% ABSTRACT
%%%%%%%%%%%%%%%%%%%%%%%%%%%%%%%%%%%%%%%%%%%Style Guild
\begin{abstract}
% In this paper, we propose a single-stage panoptic segmentation framework that focuses on deployment speed-up for real-time applications. % talk about how this is not achievable with current methods
% Our method alleviates the computational complexity associated with the instance segmentation sub-task by reusing the information inferred from object detection and semantic segmentation. % some quick explanation about how we achieve this
% The resulting network contains a single-stage data flow that does not require feature map re-sampling, thus leaving a larger space for speeding up in deployment engine. % We said this before, here you can say that this is cleaner and produces a simpler, more effective architecture
% Our proposed method shows comparable performance with the state-of-the-art on several benchmark datasets, including Cityscapes and COCO, while enjoying a considerably faster inference speed, which makes it suitable for real-time applications with high-resolution input images.
% % TODO: update the following results.
% % More specifically, our model achieves 80ms when fed with input images of size 2048 * 1024 on a XX GPU with TensorRT. % talk about how our method is more friendly to TensorRT
% Possibly to add: Nobody else has good real time panoptic + what are our numbers? 
Panoptic segmentation is a complex full scene parsing task requiring simultaneous instance and semantic segmentation at high resolution. Current state-of-the-art approaches cannot run in real-time, and simplifying these architectures to improve efficiency severely degrades their accuracy. In this paper, we propose a new single-shot panoptic segmentation network that leverages dense detections and a global self-attention mechanism to operate in real-time with performance approaching the state of the art. We introduce a novel parameter-free mask construction method that substantially reduces computational complexity by efficiently reusing information from the object detection and semantic segmentation sub-tasks.
% camera-ready
The resulting network has a simple data flow that requires no feature map re-sampling, enabling significant hardware acceleration.
% submission
% The resulting network has a simple data flow that does not require feature map re-sampling or clustering post-processing, enabling significant hardware acceleration.
Our experiments on the Cityscapes and COCO benchmarks show that our network works at 30 FPS on $1024\times2048$ resolution, trading a $3\%$ relative performance degradation from the current state of the art for up to $440\%$ faster inference.
\end{abstract}
\let\thefootnote\relax\footnotetext{$^*$ Equal contribution. }
\let\thefootnote\relax\footnotetext{This work was done when Rui Hou was an intern at Toyota Research Institute (TRI).}
%%%%%%%%% BODY TEXT
\section{Introduction}
% Briefly panoptic segmentation
Scene understanding is the basis of many real-life applications, including autonomous driving, robotics, and image editing. Panoptic segmentation, proposed by Kirillov et al.\ \cite{kirillov2019panoptic}, aims to provide a complete 2D description of a scene. This task requires each pixel in an input image to be assigned to a semantic class (as in semantic segmentation) and each object instance to be identified and segmented (as in instance segmentation). Facilitated by the availability of several open-source datasets (e.g.\ Cityscapes \cite{cordts2016cityscapes}, COCO \cite{lin2014microsoft}, Mapillary Vistas \cite{neuhold2017mapillary}), this topic has drawn a lot of attention since it was first introduced \cite{li2018learning, kirillov2019panoptic2, xiong2019upsnet, porzi2019seamless}.

% Current status of panoptic segmentation
In panoptic segmentation, pixels are categorized in two high level classes: \textit{stuff} representing amorphous and uncountable regions (such as sky and road), and \textit{things} covering countable objects (such as persons and cars). These two categories naturally split the panoptic segmentation task into two sub-tasks, namely semantic segmentation and instance segmentation. Most recent approaches use a single backbone for feature extraction and add various branches on top of the shared representations to perform each downstream task separately, generating the final panoptic prediction with fusion heuristics~\cite{kirillov2019panoptic2, xiong2019upsnet, porzi2019seamless}.

% Short in inference speed
To date, most studies on panoptic segmentation focus on improving model accuracy, either by integrating more advanced semantic and instance segmentation methods~\cite{porzi2019seamless,li2019attention} or by introducing novel information flow and loss functions\cite{li2018learning,sofiiuk2019adaptis}. None of these methods are suitable for real-time applications due to prohibitively slow inference speeds. A small subset of recent works is making progress towards faster panoptic segmentation algorithms \cite{de2019fast, deeperlab2019} but at a significant cost in terms of accuracy.
% E.g. FPSNet~\cite{de2019fast} gains $130\%$ inference efficiency degrading $8\%$ accuracy compared to Seamless~\cite{porzi2019seamless}.

% insert first figure
\begin{figure}[t!]
\begin{center}
\hbox{\hspace{-0.6mm} \includegraphics[width=0.95\linewidth]{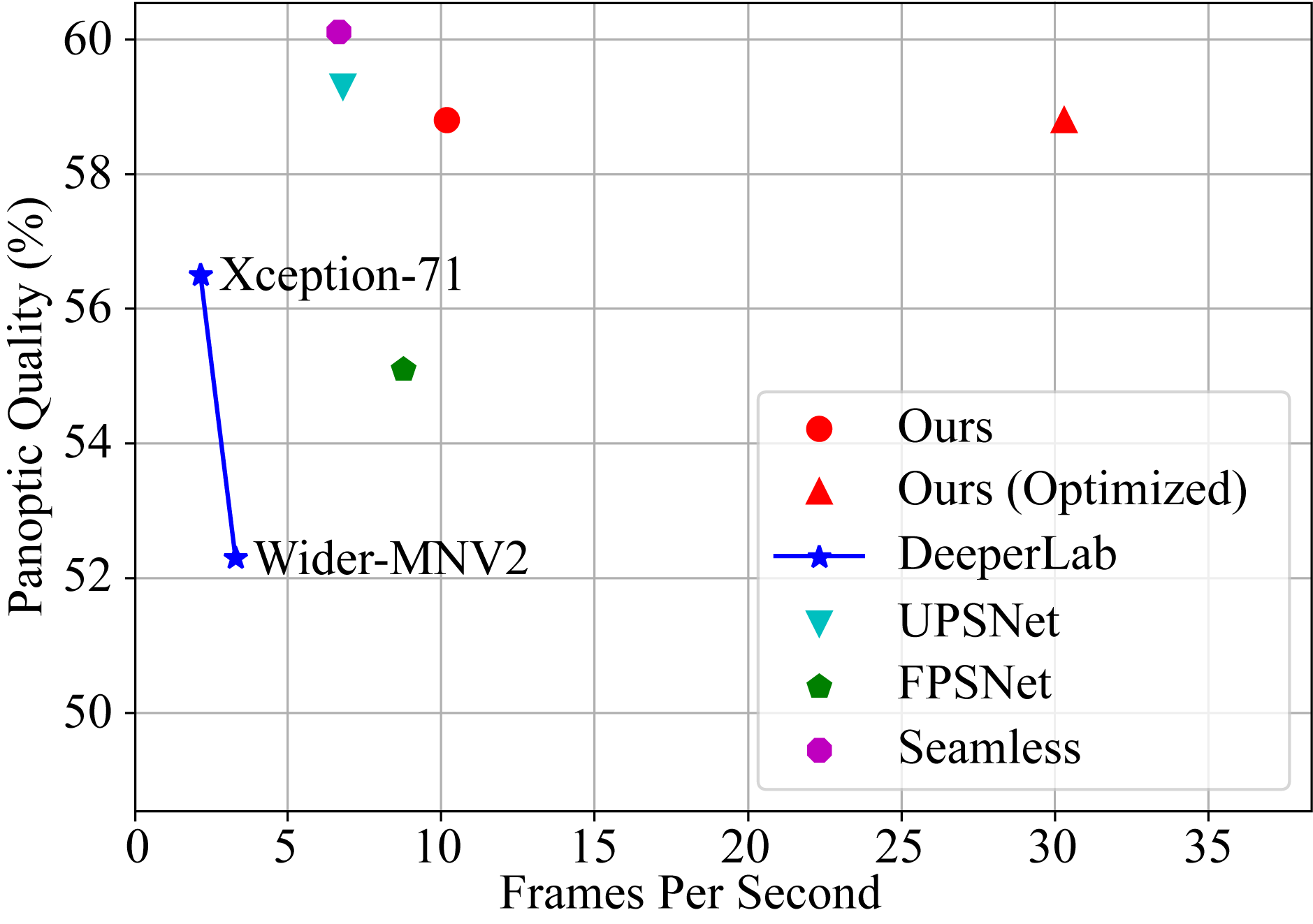}}
\end{center}
\vspace{-8mm}
   \caption{\textbf{Inference times and panoptic quality} (PQ) for the state of the art and our method on the Cityscapes validation set at $1024 \times 2048$ resolution with a ResNet-50-FPN backbone (except for DeeperLab). Our method runs in real-time at a competitive accuracy.}
\label{fig:infer_time}
\vspace{-3mm}
\end{figure}

% Introduction of our work
To achieve \emph{high quality panoptic segmentation under real-time constraints}, we identify two key opportunities for streamlining existing frameworks.
Our first observation is that most accurate instance segmentation methods follow a ``detect-then-segment" philosophy, but a significant amount of information is discarded during the ``detect" phase. Specifically, ``dense" object detection algorithms such as YOLO~\cite{redmon2018yolov3}, RetinaNet~\cite{lin2017focal} and FCOS~\cite{tian2019fcos} first generate a super-set of bounding box proposals (at least one per location), wherein multiple proposals may correspond to a single target object. Then, \ac{NMS} or an equivalent filtering process picks out predictions with the highest confidence and ignores the rest. This selection strategy discards lower ranking proposals generated by the network, even though they might have significant overlap with the ground truth. We instead propose to \textbf{reuse dense bounding box proposals discarded by NMS to recover instance masks directly}, i.e.~without re-sampling features~\cite{xiong2019upsnet} or clustering post-processing~\cite{deeperlab2019}. 

Second, we observe that semantic segmentation captures much of the same information as detection, especially in existing panoptic segmentation frameworks. For example, in \cite{xiong2019upsnet,kirillov2019panoptic2,li2018learning,porzi2019seamless}, class predictions for object detections are a subset of those for semantic segmentation, and are produced from identical representations. Hence, \textbf{sharing computations across semantic segmentation and detection streams can significantly reduce the overall complexity}.

Given these insights, we explore how to maximally reuse information in a single-shot, fully-convolutional panoptic segmentation framework that achieves real-time inference speeds while obtaining performance comparable with the state of the art. Our \textbf{main contributions} are threefold:
%\begin{itemize}[topsep=0pt,itemsep=-1ex,partopsep=1ex,parsep=1ex]
(i) we introduce a novel panoptic segmentation method extending dense object detection and semantic segmentation by reusing discarded object detection outputs via parameter-free global self-attention;
(ii) we propose a single-shot framework for real-time panoptic segmentation that achieves comparable performance with the current state of the art as depicted in Figure~\ref{fig:infer_time}, but with up to 4x faster inference;
    % We need to somehow quantify "fast" here, how much are we improving in speed and how much are we losing in performance? You can direct to Fig. 1
    %\item We propose a fast framework for real-time panoptic segmentation, with inference time X% faster than current state-of-the-art methods while maintaining Y% of its performance.
    %\item We apply a novel mask loss that enhances panoptic segmentation performance without introducing additional computation complexity at inference time. 
(iii) we provide a natural extension to our proposed method that works in a weakly supervised scenario.

% Our proposed framework also rethinks a common practice of current best performing object detection approaches, namely a redundant object proposal strategy. 

% \textit{can we make use of this discarded information to produce finer predictions?}

% % The story of reusing abandoned object detection results
% Our proposed framework also rethinks a common practice of current best performing object detection approaches, namely a redundant object proposal strategy. Algorithms such as Mask R-CNN \cite{he2017mask} and YOLO \cite{redmon2018yolov3} first generate a super-set of bounding box proposals, in which multiple proposals may correspond to a single target object. % 
% Then a \ac{NMS} or equivalent process picks out predictions with high confidence and suppresses the rest. %
% Such selection strategy eliminates low ranking proposals generated by the network, however \textit{can we make use of this discarded information to produce finer predictions?} 

% Note that in this scenario similar information is produced separately, generating unnecessary computational overhead. % What is this sentence?

\section{Related Work}
\subsection{Instance Segmentation}
\begin{figure*}[t!]
\begin{center}
\includegraphics[width=0.98\linewidth]{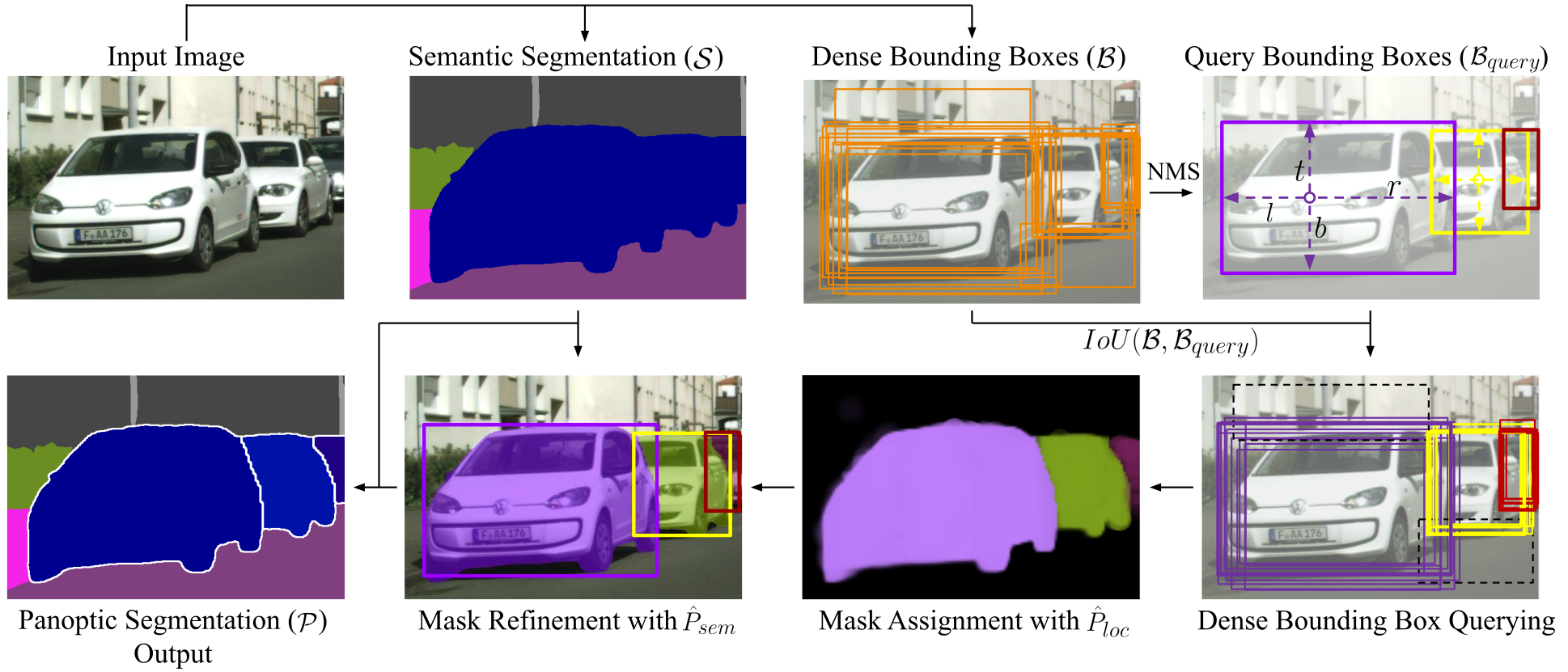}
\end{center}
\vspace{-6mm}
   \caption{\textbf{Panoptic segmentation from dense detections.}  We obtain a panoptic segmentation $\mathcal{P}$ from a semantic segmentation $\mathcal{S}$ and dense bounding box predictions $\mathcal{B}$. We first select the highest-confidence ``query" bounding boxes through NMS; then we estimate a location based mask probability $\hat{P}_{loc}$ through self-attention of the query boxes and the dense box predictions. Finally, we refine instance masks with the semantic probability map $\hat{P}_{sem}$ and produce our panoptic output.}
%   The process of generating panoptic segmentation using semantic segmentation and dense object detection predictions.}
\label{fig:general-method}
\vspace{-4mm}
\end{figure*}
%leveraging a parameter-free mask reconstruction algorithm using bounding box self-attention

% two-stage instance segmentation
Instance segmentation requires distinct object instances in images to be localized and segmented. Recent works can be categorized into two types: two-stage and single-stage methods. 
Represented by Mask R-CNN \cite{he2017mask} and its variations \cite{liu2018path, chen2019hybrid, huang2019mask}, two-stage algorithms currently claim the state of the art in accuracy. 
The first stage proposes a set of regions of interest (RoIs) and the second predicts instance masks from features extracted using RoIAlign \cite{he2017mask}. 
This feature re-pooling and re-sampling operation results in large computational costs that significantly decrease efficiency, rendering two-stage models challenging to deploy in real-time systems.
% single-stage instance segmentation
Single-stage methods, on the other hand, predict instance location and shape simultaneously. 
Some single-stage methods follow the detect-then-segment approach, with additional convolutional heads attached to single-stage object detectors to predict mask shapes \cite{xu2019explicit, uhrig2018box2pix, yolact19, xie2019polarmask}. 
Others learn representations for each foreground pixel and perform pixel clustering to assemble instance masks during post-processing~\cite{neven2019instance, gao2019ssap, de2017semantic, newell2017associative, liang2017proposal}. 
The final representation can be either explicit instance-aware features \cite{neven2019instance, liang2017proposal}, implicitly learned embeddings~\cite{de2017semantic, newell2017associative}, or affinity maps with surrounding locations at each pixel~\cite{gao2019ssap}. 

% our work 
Following the detect-then-segment philosophy, our work tackles instance segmentation solely based on object detection predictions. In this sense, it is similar to works which densely predict location-related information to separate different instances~\cite{neven2019instance, deeperlab2019, de2017semantic}.
% We probably don't want to include this paper.
% For example, Brabandere et al. proposes to learn implicit per-pixel embedding using a discriminative loss \cite{de2017semantic}. However, this work ignore the location related features and rely on clustering algorithms to recover the proposal location, which is time consuming and cannot be scaled with the increased number of objects. 
% For example, Neven et al. \cite{neven2019instance} propose to regress offset vectors for each pixel relative to its corresponding instance center, which can be seen as learning a spatial embedding to capture both location and shape information. 
However, even though these methods claim real-time performance, their location proposal and pixel clustering processes are formulated in an iterative pixel-based manner, which makes inference time dramatically increase with the number of instances. Our framework regresses bounding boxes in such a way that proposals can be selected directly through NMS.

Our framework is also similar to the Proposal-Free Network (PFN) \cite{liang2017proposal}, as we use instance bounding box coordinates as features to group same-instance pixels. However, PFN relies on predicting the number of instances and on a clustering algorithm which is sensitive to parameter tuning. We use bounding boxes as both proposals and embeddings, and apply a straightforward correlation matrix-based approach for mask recovery. Casting instance segmentation as an object detection problem dramatically decreases the number of model parameters and hyper-parameters involved in training.

\subsection{Panoptic Segmentation}
% introduce panoptic segmentation
Originally proposed in \cite{kirillov2019panoptic}, panoptic segmentation has been widely accepted by the computer vision community as a major task for dense visual scene understanding. 
Each pixel in an image needs to be assigned a semantic label as well as a unique identifier (ID) if it is an object instance. 
Pixels with the same label belong to the same category, and pixels with the same ID (if present) furthermore belong to the same object instance.

Current state-of-the-art panoptic segmentation work uses a multi-decoder network to perform prediction with redundant information, as well as more sophisticated instance or semantic segmentation heads for better performance \cite{xiong2019upsnet, kirillov2019panoptic2, li2018learning, liu2019end, porzi2019seamless, li2019attention}.

While obtaining state-of-the-art accuracy on panoptic segmentation, these methods are far from real-time applications mainly due to two reasons: 1) the instance segmentation branch contains two stages, which makes it difficult to be accelerated by inference engines (e.g.\ TensorRT~\cite{tensorrt}); and 2) similar information (i.e., semantics) is processed by both branches, demanding redundant kernel operations that slow down the entire network. %

% We might want to skip the following part as they are old studies and not very related
% Some earlier works tackling the panoptic segmentation problem using multi-task joint networks [2][3][4] that predict semantic segmentation and instance segmentation embedding with separate decoders with no redundant or overlapping information between individual sub-networks. However, quantitative results in these works also indicate that these types of methods underperform the state-of-the-art. The post-processing step to decode the embedding into direct predictions is also another problem with these methods as they are often computationally expensive and hard to scale with the number of instances in an image.

Recently, DeeperLab \cite{deeperlab2019} proposed to solve panoptic segmentation with a one-stage instance parser without using the same design as Mask R-CNN \cite{he2017mask}. However, this approach to instance segmentation involves prediction of key points and instance center offsets with a complicated post-processing step, which prevents it from running in real-time despite its single-shot design. Some other methods \cite{de2019fast, weber2019singleshot} focus on pushing inference time to real-time by using simpler instance separation methods, however their accuracy is not comparable with the state of the art.

In this work, we resolve the deployment difficulties associated with multi-stage models by grafting a single-stage panoptic head directly onto the backbone, which simultaneously predicts instance as well as semantic segmentation information. Our method is more compatible with inference engine deployment, requiring less memory copy and resampling. Finally, our method also makes efficient usage of all network kernels, as no redundant information is predicted from different sequential branches. As a result, we can achieve real-time inference with accuracy comparable with the current state of the art, thus setting a new baseline for real-time panoptic segmentation methods.

% In this section, we propose a novel formulation of the panoptic segmentation and justify our argument that dense bounding box regression and semantic segmentation is sufficient to address panoptic segmentation problem. 
% We first introduce the idea of utilizing dense bounding box predictions to generate instance masks, which is one of the main novelties of our work.
%In this work, we argue that fine instance masks can be extracted from densely predicted object bounding boxes. Our proposed method can be built upon any single-stage object detectors such as YOLO \cite{redmon2018yolov3}, RetinaNet \cite{lin2017focal} and FCOS \cite{tian2019fcos}. Similar to regular single-stage detectors, the predictions with high confidence are picked out through NMS from all predictions. We call these predictions \textit{query} bounding boxes and assume each of them represents the location of an object instance. Even the rest of the bounding box predictions are not optimal, they still represent approximate locations of the corresponding object that a specific location in the feature map belongs to. Using the full set of  bounding box predictions 
\section{{Panoptic from Dense Detections}}
\label{sec:panoptic_transforming}

\subsection{{Problem Formulation}}

%%% Problem statement and methodology: describe in general how we can solve panoptic segmentation from box.
In this section, we describe our approach to address the panoptic segmentation task ($\mathcal{P}$) by solving a semantic segmentation task ($\mathcal{S}$) and a dense bounding box detection task ($\mathcal{B}$) as depicted in Figure~\ref{fig:general-method}.
%We first denotes the target predictions of these three tasks.
%
The objective of \textbf{panoptic segmentation} is to predict semantic and instance IDs for each pixel $(x,y)$ in the input image:
% \begin{equation}
%     \mathcal{P}^t(x,y) = (c,k),\quad c\in \{1, ..., N\}, k\in \mathbb{Z},
% \end{equation}
% camera-ready Z -> N
\vspace{-2mm}
\begin{equation}
    \mathcal{P}(x,y) = (c,k),\quad c\in \{1, ..., N\}, k\in \mathbb{N},
    \vspace{-2mm}
\end{equation}
where $c$ is the semantic class ID, $k$ is the instance ID (with $0$ for all \textit{stuff} classes) and $N$ is the total number of classes, including \textit{stuff} ($N_{\textnormal{stuff}}$) and \textit{things} ($N_{\textnormal{things}}$) classes.
% -----
% camera-ready
In the \textbf{semantic segmentation}  sub-task, we predict a distribution over semantic classes for each pixel $(x,y)$, $ \mathcal{S}(x,y)$.
We denote $\hat{P}_{sem}(x,y,c)$ as the predicted probability at pixel $(x,y)$ for semantic class $c$, given by:
\vspace{-2mm}
\begin{equation}
  \hat{P}_{sem}(x,y,c) =  \mathcal{S}(x,y)[c], \quad \mathcal{S}(x,y)\in \mathbb{R}^N 
  \label{eq:predicted_sem_prob}
  \vspace{-2mm}
\end{equation}
% submission
% In the \textbf{semantic segmentation}  sub-task, we predict a distribution over semantic classes for each pixel $(x,y)$:
% \begin{equation}
%     \mathcal{S}(x,y)=\textbf{s},\quad \textbf{s}\in \mathbb{R}^N, {\textbf{s}}[c] = \hat{P}_{sem}(x,y,c),
% \end{equation}
% where the highest probability over the estimated pixel-wise distribution is taken as the final prediction.
% -----
% \begin{equation}
%     \mathcal{S} = \argmax_\textnormal{c}{\mathcal{S}_{c}(x,y)}
% \end{equation}
%
%In the task of \textbf{Dense bounding box predictions} are used by one-stage object detectors to locate and classify target objects in an input image, which are represented by a set of ground-truth bounding boxes defined as:
In the \textbf{dense bounding box detection} sub-task, we predict at least one bounding box at each image pixel:
% \begin{equation}
% \begin{gathered}
%     \mathcal{B}^t = \{B_i^t\},\quad B_i^t=(\textbf{b}^t_i,c_i^t), \\
%     \textbf{b}_i^t=(x_1^i,x_2^i,y_1^i,y_2^i) \in \mathbb{R}^4,\quad c_i^t \in \{1, ..., N_{thing}\}
% \end{gathered}
% \end{equation}
\vspace{-2mm}
\begin{equation}
\begin{gathered}
    \mathcal{B}(x,y) = \textbf{B},\quad \textbf{B}=(\textbf{b},c),\\
    \textbf{b}=(x_1,x_2,y_1,y_2) \in \mathbb{R}^4,\quad c \in \{1,...,{N_{\textnormal{things}}\}},
\end{gathered}
\label{eq:box}
\vspace{-3mm}
\end{equation}
% camera-ready: left-top -> top-left
where $(x_1, y_1)$  and $(x_2, y_2)$ are the coordinates of the top-left and bottom-right corners of bounding box $\textbf{B}$ that pixel $(x,y)$ belongs to;
%$c$ is the estimated probability of objectness with respect to each \textit{thing} classes.
$c$ is the predicted class ID for the corresponding bounding box.
We note that because $\mathcal{S}$ and $\mathcal{B}$ are of fixed dimensions, they can be directly learned and predicted by fully convolutional networks.

%of target class out of all $N_{thing}$ possible object classes in the dataset.
%To generate the final targets we have dense predictions as 
% \begin{equation}
%     \mathcal{B}^p(x,y) = B^p,\quad B^p=(\textbf{b}^p,P^p(c))
% \end{equation}
%where the definition of $\textbf{b}^p$ is similar to that of $\textbf{b}_i^t$; $P^p(c)$ represents the likelihood (ranging from 0 to 1) that this bounding box is of a specific class $c$ and it is usually a product of multiple probabilities such as class probability, objectiveness, etc.
\subsection{Parameter-Free Mask Construction}
Given $\mathcal{S}$ and $\mathcal{B}$, we introduce a parameter-free mask reconstruction algorithm to produce instance masks based on a global self-attention mechanism.
%we can convert them into a dense panoptic prediction through our novel parameter-free mask reconstruction algorithm. 
% box proposal
%In this work we call the entire set of dense bounding box predictions, $\mathcal{B}^p$, \textit{key} bounding boxes. 
Note that the following operations are embarrassingly parallel.
We first obtain a reduced set of bounding box proposals from $\mathcal{B}$ through \ac{NMS}:
%process on $\mathcal{B}^p$, we can obtain a set of local maximum proposals after ranking and evaluating their $P^p(c)$ and \ac{IoU}. In this work, we call these proposals \textit{query} bounding boxes, ${\textbf{B}}_{query}$, as we will later use them to \say{search} instance masks:
% \begin{equation}
%     \textbf{B}_{query}=\{B_j^p\}, \quad B_j^p=(\textbf{b}_j^p ,c_j).
% \end{equation}
% ${\textbf{B}}_{query}$, as we will later use them to \say{search} instance masks:
\vspace{-2mm}
\begin{equation}
\begin{gathered}
    \mathcal{B}_{query}=\{\textbf{B}_j\}, \quad \textbf{B}_j=(\textbf{b}_j ,c_j).
\end{gathered}
\end{equation}
%where $c_j$ is the most likely class of the selected bounding box.
We denote the proposal set as ${\mathcal{B}}_{query}$ because we will use them to \say{search} for instance masks.
%
% Next, we construct one instance mask $\mathcal{M}_j$ for each $\textbf{B}_j$ in $\textbf{B}_{query}$. All pixels in the image can contribute to $\mathcal{M}_j$ (and not just those contained in $\textbf{b}_j$) as follows:
% \begin{equation}
%     \mathcal{M}_j(x,y)=
%     % \Bigg\{ \begin{matrix}
%     % k, if P_loc \and P_sem \geq \sigma
%     % \\ -1
%     % \end{matrix}
%     \begin{cases}{}
%   1,\quad  \textbf{P}^j_{loc} \cap \textbf{P}^j_{sem} (j)\geq \sigma
%     \\
%     0, \quad \text{else}
%     \end{cases}
% \label{eq:mask}
% \end{equation}{}
For each query box $\textbf{B}_j$, we construct a global mask probability map given by:
\vspace{-2mm}
\begin{equation}
\begin{gathered}
\mathcal{M}(x,y,j) = \hat{P}_{loc}(x,y,j)\cdot \hat{P}_{sem}(x,y,c_j),
\label{eq:mask_map}
\end{gathered}
\vspace{-2mm}
\end{equation}
% \noindent where $\mathcal{M}(x,y)=j$ indicates that pixel $(x,y)$ belongs to object $j$ which is represented by a \textit{query} bounding box $B_j^p$ and $-1$ indicates that this pixel is stuff.
%$\sigma$ is a heuristic likelihood threshold to separate background and foreground pixels for a specific instance.
% -----
% camera-ready
where $\hat{P}_{loc}(x,y,j)$ is an estimated probability that pixel $(x,y)$ is inside object $j$'s bounding box. 
We estimate this probability by self-attention between the global set of boxes $\mathcal{B}$ and the query box $\textbf{B}_j$ with Intersection over Union (IoU):
\vspace{-2mm}
\begin{equation}
\label{eq:ploc}
    \hat{P}_{loc}(x,y,j) = \IoU{(\mathcal{B}(x,y),\textbf{B}_j)},
    \vspace{-1mm}
\end{equation}
where $\IoU(\textbf{B}_i, \textbf{B}_j) = \text{intersection}(\textbf{b}_i, \textbf{b}_j) / \text{union}(\textbf{b}_i, \textbf{b}_j)$, and $\hat{P}_{sem}(x,y,c_j)$ is the predicted probability that pixel $(x,y)$ shares the same semantic class $c_j$ given by Eq.~\ref{eq:predicted_sem_prob}.
% submission
% where $\hat{P}_{loc}(x,y,j)$ is an estimated probability that pixel $(x,y)$ shares the same bounding box as object $j$. 
% We approximate this probability with self-attention between the global set of boxes $\mathcal{B}$ and the query box $\textbf{B}_j$:
% \begin{equation}
% \label{eq:ploc}
%     \hat{P}_{loc}(x,y,j) = \IoU{(\mathcal{B}(x,y),\textbf{B}_j)}.
%     \vspace{-2mm}
% \end{equation}{}
% where the highest probability over the estimated pixel-wise distribution is taken as the final prediction.

% $\hat{P}_{sem}(x,y,c_j)$ is the probability that pixel $(x,y)$ shares the same semantic class as object $j$, which is given by:
% \begin{equation}
%     \hat{P}_{sem}(x,y,c_j) = \mathcal{S}(x,y)[c_j].
%     % \hat{P}_{sem}(x,y,j) = \mathcal{S}(x,y)[c_j].
% \end{equation}
% -----
To construct the final instance masks $\{\textbf{M}_j\}$, we apply a simple threshold $\sigma$ to the global mask probability map:
\vspace{-2mm}
\begin{equation}
    \textbf{M}_j(x,y) = \mathcal{M}(x,y,j)>\sigma.
    \vspace{-2mm}
\end{equation}

To produce a final panoptic segmentation, we follow the conventional fusion strategy in~\cite{kirillov2019panoptic}. A graphical illustration of the complete method is shown in Figure~\ref{fig:general-method}.
% In the end, panoptic segmentation is determined by both dense semantic and instance association:
% \begin{equation}
%     \mathcal{P}(x,y) = 
%     \begin{cases}{}
%   ( c_{\argmax \mathcal{M}(x,y,j)}, \argmax \mathcal{M}(x,y,j)),\quad \text{things}
%     \\
%   (\argmax_c \mathcal{S}(x,y), 0),\quad \text{stuff}
%     \end{cases}
% \end{equation}
% We name the above panoptic segmentation generation process as \textbf{Panoptic Transforming} and it is also illustrated in Fig.~\ref{fig:general-method}.
%%% Now we provide light discussion on why this solution is light.
%n this approach, the only necessary predictions are semantic logits ($\mathcal{S}$) and dense bounding box detection ($\mathcal{B}$). No additional parameter is needed for instance segmentation as detection results are comprehensively used in \textbf{both} candidate proposal and bounding box self-attention.
%We concrete this idea into a novel real-time single-stage architecture achieving near state-of-the-art performance, introduced in the following section.
We demonstrate the efficacy of our proposed method in a novel real-time single-stage architecture in the following section.
We note, however, that this method can be generalized to any architecture that provides predictions of $\mathcal{B}$ and $\mathcal{S}$.

% A series of \textit{query} bounding boxes are predicted with each representing the location of a distinct object instance. At the same time, the framework also generates a dense map of \textit{key} bounding boxes with each location representing the corresponding object instance it belongs to. Given a predicted query bounding box, in order to assemble its instance mask a natural step is to collect all locations that belongs to the same instance in the key bounding box map. This instance mask assembly process is tensorizable and thus highly parallelizable and fast. We also integrate semantic segmentation predictions into this process to enhance the concordance between semantic segmentation and instance segmentation.

% \section{Panoptic Transformer Network}
\section{Real-Time Panoptic Segmentation Network}
\begin{figure*}[t]
\begin{center}
 \includegraphics[width=1\linewidth]{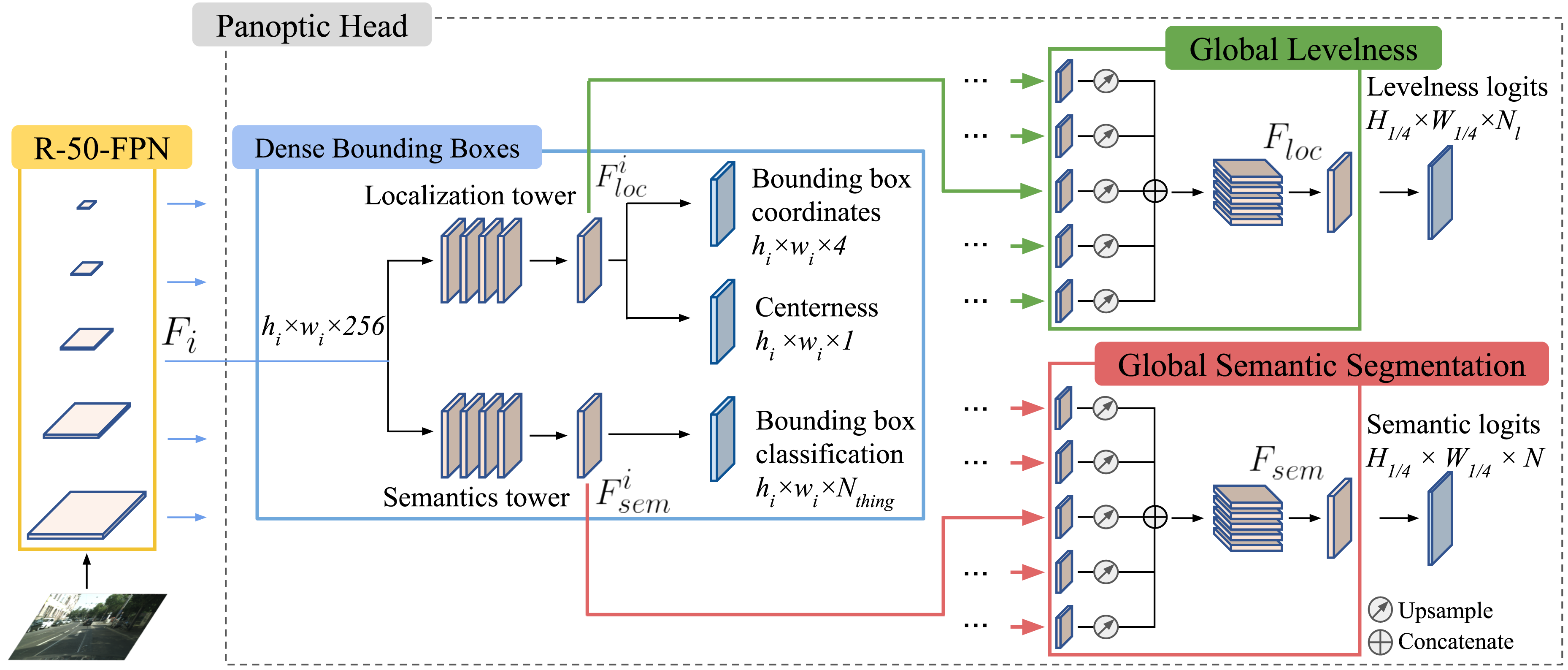}
% \fbox{\rule{0pt}{2in} \rule{.9\linewidth}{0pt}}
\end{center}
\vspace{-5mm}
\caption{\textbf{Real-time panoptic segmentation.} Our model uses ResNet-50-FPN as backbone. The multi-scale feature maps are fed into a \textit{unified} panoptic head. We predict dense bounding boxes at each FPN Level. We also upsample and concatenate intermediate feature maps across all levels to predict global levelness and semantic logits.}
\label{fig:flow_chart}
\vspace{-3mm}
\end{figure*}
We propose a single-stage panoptic segmentation network capable of real-time inference, as shown in Figure~\ref{fig:flow_chart}.
As in other recent works on panoptic segmentation~\cite{porzi2019seamless,kirillov2019panoptic2}, our architecture is built on a ResNet-style~\cite{he2016deep} encoder with a Feature Pyramid Network (FPN) \cite{lin2017feature}. Our FPN module consists of 5 levels corresponding to strides of 128, 64, 32, 16 and 8 with respect to the input image.

Our panoptic segmentation architecture is inspired by FCOS~\cite{tian2019fcos}, a fast and accurate fully convolutional, anchor-less object detector. However, we use a finer-grained target assignment scheme and additional single-convolution layers for semantic segmentation and a novel ``levelness" prediction, which will be described later in Section~\ref{sec:target}.
Our framework also leverages a fully tensorizable mask construction algorithm described in Section~\ref{sec:panoptic_transforming}, and we propose an explicit instance mask loss which further improves the quality of the final panoptic segmentation. 
\vspace{-2mm}
\subsection{Target Assignment\label{sec:target}}
\vspace{-2mm}
We formulate object detection as a per-pixel prediction problem. Specifically, let $\textbf{F}_{i} \in \mathbb{R}^{h_i \times w_i \times C}$ be the feature map at layer $i$ of the FPN, with stride $z$. We assign a target to each location $(x,y)$ on $\textbf{F}_{i}$ for all $i$. 
Since each $(x,y)$ is the center of a receptive field, we can recover the pixel location in the original image using the relation: $(x_o, y_o) = ({\lfloor\frac{z}{2}\rfloor+ xz}, \lfloor\frac{z}{2}\rfloor + yz)$.
%
% Differing from FCOS, which uses only bounding boxes in its assignment scheme, w
% -----
% camera-ready
If a pixel location $(x_o,y_o)$ falls within one of the ground truth instance masks $\mathcal{M}^t:\{\textbf{M}^t_j\}$, then we consider it a foreground sample. Since instance masks are non-overlapping, location $(x_o,y_o)$ is associated with \textit{only one} mask $\textbf{M}^t_j$ and its corresponding bounding box $\textbf{B}^t_j$. This avoids the ``ambiguous pixel" issue described in \cite{tian2019fcos}.
If location $(x_o, y_o)$ is associated with ground truth $\textbf{B}^t_j$, then we assign the following regression offsets $\textbf{t}^t_{xy} = (l, t, r, b)$ to it: 
% submission
% If a pixel location $(x_o,y_o)$ falls within one of the ground truth instance masks $\mathcal{M}_{GT}:\{ \textbf{M}_{GT_j}\}$, then we consider it a foreground sample. Since instance masks are non-overlapping, location $(x_o,y_o)$ is associated with \textit{only one} mask $\textbf{M}_{GT_j}$ and its corresponding bounding box $\textbf{B}_{GT_j}$. This avoids the ``ambiguous pixel" issue described in \cite{tian2019fcos}.
% If location $(x_o, y_o)$ is associated with ground truth $\textbf{B}_{GT_j}$, then we assign the following regression offsets $\textbf{t}_{xy} = (l, t, r, b)$ to it: 
% -----
\vspace{-3mm}
\begin{equation}
\label{eq:bbox_regression}
l = x_o -x_1, \hspace{3 pt} t = y_o-y_1,\hspace{3 pt}
r = x_2 - x_o, \hspace{3 pt} b = y_2 - y_o,
\end{equation}
% -----
% camera-ready
where $\textbf{b}^t_j=(x_1,y_1,x_2,y_2)$ are the bounding box coordinates as defined in Eq~\ref{eq:box}. These 4-directional offsets are also showcased in Figure~\ref{fig:general-method}. We define $\mathcal{T}_{i}$ to be the set of regression targets \{$\textbf{t}^t_{xy}$\} assigned to feature map $F_{i}$.
% submission 
% where $\textbf{b}_{GT_j}=(x_1,y_1,x_2,y_2)$ are the bounding box coordinates as defined in Eq~\ref{eq:box}. These 4-directional offsets are also showcased in Figure~\ref{fig:general-method}. We define $\mathcal{T}_{i}$ to be the set of regression targets \{$\textbf{t}_{xy}$\} assigned to feature map $F_{i}$.
% -----

Since it is possible that locations on multiple FPN levels $\textbf{F}_{i}$ resolve to the same $(x_o, y_o)$, we disambiguate them by removing offset $\textbf{t}^t_{xy}$ from level $i$ if it does not satisfy the following policy:
\vspace{-2mm}
\begin{equation}
   \textbf{t}^t_{xy}\in \mathcal{T}_{i},\text{iff}\quad m_{i-1} <= \max(l, t, r, b) <= m_i,
    \label{eq:assignment}
    \vspace{-2mm}
\end{equation}
% camera-ready: added "manually defined".
where ${m_i}$ is the heuristic value of maximum object size for target assignment in $\textbf{F}_i$.
At each location, we also predict centerness, $o^t_{xy}$~\cite{tian2019fcos}:
\vspace{-3mm}
\begin{equation}
\label{eq:centerness}
    o^t_{xy} = \sqrt{\frac{\min(l,r)}{\max(l,r)} \times \frac{\min(t,b)}{\max(t,b)}}.
    \vspace{-2mm}
\end{equation}
% ----- 
% camera-ready
During \ac{NMS}, we multiply the bounding box confidence with predicted centerness to down-weight bounding boxes predicted near object boundaries. 
For each location, we also predict the object class, $c^t_{xy}$, of the assigned bounding box $\textbf{B}^t_j$. 
Thus, we have a $6$-dimensional label, $(\textbf{t}^t_{xy}, c^t_{xy}, o^t_{xy})$, at each foreground location $(x,y)$ on each $\textbf{F}_{i}$.
% submission
% Centerness down-weights bounding boxes predicted near object boundaries during \ac{NMS}. 
% For each location, we also predict the semantic class of the assigned  bounding box $\textbf{B}_{GT_j}$. Thus, we have a $6$-dimensional label, $(\textbf{t}_{xy}, c_{xy}, o_{xy})$, at each foreground location $(x,y)$ on each $\textbf{F}_{i}$.
% -----

%We additionally assign the semantic class $c_{GT_j}$ and feature map level $i$ as targets for each foreground pixel location in the original image.
%For locations assigned to background, we set $c_{GT} = 0$ and $i=0$.
%%%%%%%%%%%%%%%%%%%%%%%%%%%%%%%%%%%%%%%%%%%%%%%%%%%%%%%%%%%%%%%%%%%%%%%%%%%%%%%%%%%%%%%%%%%%%%%%%%%%%%%%%%%%%%%%%%%%%%%%%%%%%%%%%%%%%%%%%%%%%%%
\subsection{Unified Panoptic Head}
\label{sec:panoptic_head}
% We design a unified panoptic head which is shared by each of the multi-scale feature maps produced by our backbone.
We design a unified panoptic head predicting both semantic segmentation and dense bounding boxes from the multi-scale features maps. 
%
% On each feature map, we apply two feature towers, one for localization and the other for semantics, as shown in Figure~\ref{fig:flow_chart}. 
% Each tower contains 4 sequential convolutional blocks (Conv + GroupNorm + ReLU).

\textbf{Per Level Predictions}
At each location $(x,y)$ of FPN level $\textbf{F}_i$, we predict dense bounding boxes using two feature towers, one for localization and the other for semantics, as shown in Figure~\ref{fig:flow_chart}. Each tower contains 4 sequential convolutional blocks (Conv + GroupNorm + ReLU). The towers are shared across different FPN levels.
We directly predict bounding box offsets $\hat{\textbf{t}}_{xy}$, centerness $\hat{o}_{xy}$, and bounding box class probabilities $\hat{\textbf{c}}_{xy}$. The bounding box offset is predicted from the localization tower and the box class probability distribution is predicted from the semantics tower.
%% -----
We adopt an $\mathrm{IoU}$ Loss (\cite{yu2016unitbox}) on bounding box regression:
% camera-ready: define parameters and remove \sum_i
\vspace{-2mm}
\begin{equation}
\mathcal{L}_{box\_reg} = \frac{1}{N_{\textbf{fg}}} \sum_{xy} \mathrm{L}_{\IoU}(\hat{\textbf{b}}_{xy}, \textbf{b}^t_{xy})\mathbb{1}_{\text{\textbf{fg}}}(x,y),
\label{eq:l_box_reg}
\vspace{-3mm}
\end{equation}
where $N_{\textbf{fg}}$ is the number of foreground (i.e. things) pixels according to the ground truth mask, $\hat{\textbf{b}}_{xy}$ and $\textbf{b}^t_{xy}$ are the absolute bounding box coordinates computed from Eq.~\ref{eq:bbox_regression} using $\hat{\textbf{t}}_{xy}$ and $\textbf{t}^t_{xy}$, and $\mathbb{1}_{\text{\textbf{fg}}}(x,y)$ is an indicator function yielding 1 when $(x,y)$ corresponds to a foreground.
% submission
% To compute a loss on bounding box regression, we adopt an $\mathrm{IoU}$ Loss (\cite{yu2016unitbox}):
% \begin{equation}
% \mathcal{L}_{box\_reg} = \frac{1}{N_{\textbf{fg}}}\sum_i \sum_{xy} \mathrm{L}_{\IoU}(\hat{\textbf{b}}_{xy}, \textbf{b}_{xy})\mathbb{1}_{\text{\textbf{fg}}}(x,y),
% \label{eq:l_box_reg}
% \vspace{-3mm}
% \end{equation}
% where $\hat{\textbf{b}}_{xy}$ and $\textbf{b}_{xy}$ are the absolute bounding box coordinates computed by inverting Eq.~\ref{eq:bbox_regression} using $\hat{\textbf{t}}_{xy}$ and $\textbf{t}_{xy}$.
% -----

We compute a loss on predicted centerness at the same locations $(x,y)$ of each FPN level $\textbf{F}_i$ using a Binary Cross Entropy (BCE):
% -----
% camera-ready: remove \sum_i
\vspace{-2mm}
\begin{equation}
\mathcal{L}_{center}  = \frac{1}{N_{\textbf{fg}}} \sum_{xy} \mathrm{L}_{\mathrm{BCE}}(\hat{{o}}_{xy}, {o}^t_{xy})\mathbb{1}_{\text{\textbf{fg}}}(x,y).\label{eq:l_center}
\vspace{-2mm}
\end{equation}
% submission
% \begin{equation}
% \mathcal{L}_{center}  = \frac{1}{N_{\textbf{fg}}}\sum_i \sum_{xy} \mathrm{L}_{\mathrm{BCE}}(\hat{{o}}_{xy}, {o}_{xy})\mathbb{1}_{\text{\textbf{fg}}}(x,y).\label{eq:l_center}
% \vspace{-2mm}
% \end{equation}
% -----
%
Finally, we predict a probability distribution over object classes $\hat{\textbf{c}}_{xy} \in \mathbb{R}^{N_{\textnormal{things}}}$ for all feature locations $(x,y)$ including background pixels. For our box classification loss $\mathcal{L}_{box\_cls}$, we use a sigmoid focal loss as in \cite{lin2017focal, tian2019fcos}, averaged over the total number of locations across all FPN levels.
% \begin{equation}
% \mathcal{L}_{box\_cls} = \frac{1}{N_{loc}}\sum_{x,y,j}{-\alpha_t(1-\hat{\textbf{c}}_{xy}[j])^\gamma \log(P^},
% \end{equation}

% \subsubsection*{Bounding box class prediction}

%%%%%%%%%%%%%%%%%%%%%%%%%%%%%%%%%%%%%%% levelness regression
\textbf{Global Predictions.}
In addition to the per level predictions, we leverage the intermediate features from the two towers ($F_{loc}^i$ and $F_{sem}^i$) to globally predict: \vspace{-2mm}
\begin{enumerate}
    \item Levelness $\mathcal{I}$: the FPN level that the bounding box at each location $(x,y)$ belongs to ($N_{l}$= the number of FPN levels +1 logits for each location with $0$ reserved for background pixels). \vspace{-2mm}
    \item Semantic segmentation $\mathcal{S}$: the semantic class probability distribution over $N$ classes. \vspace{-2mm}

\end{enumerate}
As depicted in Figure~\ref{fig:flow_chart}, we upsample each $\textbf{F}_{loc}^i$ and $\textbf{F}_{sem}^i$ to an intermediate size of $(H/4,W/4)$ and concatenate them into a global $\textbf{F}_{loc}$ and $\textbf{F}_{sem}$.
% We name this $6D$ prediction as levelness, which provides a probability estimation on which scale level that a corresponding pixel belongs to according to the assignment policy defined in Eq~\ref{eq:assignment}.
%Apart from 4D bounding box and 1D center-ness, within the localization tower, we also predict a 6D levelness map of 1/4 size of the input image to assemble the \textit{key} bounding box map for instance mask recovery. 
%We treat the prediction of levelness as a classification problem: given a location $(x, y)$ in the levelness map its class indicates which scale level its bounding box prediction belongs to, according to the assignment policy we described previously.
% As we have 5 scale levels from FPN, we use $0$ to represent background and $1 \sim 5$ to represent level indices starting from the coarsest level.
%To predict a single levelness map, we up-sample feature map at each scale level to the desired resolution bi-linearly, sum them up together, and use an additional convolutional head for prediction.
The levelness is predicted from $\textbf{F}_{loc}$ through a single convolutional layer and is supervised by the FPN level assignment policy defined in \eqref{eq:assignment}.
The levelness is trained using a multi-class cross-entropy loss:
% \begin{equation}
% \mathcal{L}_{levelness} = \frac{-1}{N_{map}}\sum_{x,y}{\sum_{i=0}^{6}{l_i^t(x,y) \log(l_i^p(x,y))}},
% \end{equation}
\vspace{-2mm}
\begin{equation}
\label{eq:levelness}
\mathcal{L}_{levelness} = \mathrm{L}_{\mathrm{CE}}(\mathcal{I},\mathcal{I}^t).
\vspace{-1mm}
\end{equation}
At inference time, for every $(x,y)$ we have one bounding box prediction $\hat{\textbf{b}}_{xy}^{(i)}$ coming from each FPN level $\textbf{F}_i$. Levelness tells us which $\hat{\textbf{b}}_{xy}^{(i)}$ to include in our global set of dense bounding box predictions $\mathcal{B}$:
\vspace{-2mm}
\begin{equation}
    % \mathcal{B}(x,y) = \mathcal{B}^{\argmax Lv(x,y)}(x,y)
    % \mathcal{B}(x,y) = \mathcal{B}[\argmax \mathcal{L}(x,y)](x,y)
    \mathcal{B}(x,y) = \hat{\textbf{b}}_{xy}^{(\argmax \mathcal{I}(x,y))}.
    \vspace{-1mm}
\end{equation}

Instead of using a separate branch for semantic segmentation \cite{li2018learning, xiong2019upsnet, kirillov2019panoptic2}, we reuse the same features as bounding box classification. Doing so dramatically reduces the number of parameters and inference time of the network. 
We predict the full class semantic logits from $F_{sem}$, which we supervise using a cross-entropy loss:
\vspace{-2mm}
\begin{equation}
% \mathcal{L}_{semantics} = \frac{-1}{N_{map}}\sum_{x,y}{\sum_{i=0}^{N-1}{c_i^t\log(c_i^p)}}
\mathcal{L}_{semantics} =  \mathrm{L}_{\mathrm{CE}}(\mathcal{S},\mathcal{S}^t),
\vspace{-2mm}
\end{equation}
where $\mathcal{S}^t$ denotes semantic labels.
We bootstrap this loss to only penalize the worst $30\%$ of predictions as in ~\cite{wu2016bridging, pohlen2017full}.
%The final semantic segmentation is generated by another iteration of up-sampling process.
%%%%%%%%%%%%%%%%%%%%%%%%%%%%%%%%%%%%%%%%%%%%%%%%%%%%%%%%%%%% mask loss
\subsection{Explicit Mask Loss}
\label{sec:mask_loss}

As discussed in~\cite{tian2019fcos}, the quality of bounding box prediction tends to drop with distance from the object center. This hurts the performance of our mask construction near boundaries. In order to refine instance masks, we introduce a loss that aims to reduce False Positive (FP) and False Negative (FN) pixel counts in predicted masks:
%on boundaries and also balance the weight across different object instances, we introduce a mask loss that directly optimizing the expectation of reconstructed mask:
\vspace{-2mm}
\begin{equation}
% \mathcal{L}_{mask} = \frac{1}{M} \sum_{j}^{|\mathcal{B}_{query}|} \frac{\beta_j}{N_j} (E_{\text{FP}_j} + E_{\text{FN}_j}),
\mathcal{L}_{mask} = \frac{1}{|\mathcal{B}_{query}|} \sum_{j}^{|\mathcal{B}_{query}|} \frac{\beta_j}{N_j} (E_{\text{FP}_j} + E_{\text{FN}_j}),
%\sum_j{\beta_j \frac{1}{K_j} \sum_{k=1}^{K_j}{((1-p_k)t_k+ \\ \max(0, p_k-\alpha)(1-t_k))}}
\vspace{-3mm}
\end{equation}
% ----
% camera-ready
where $\beta_j$ is the IoU between proposal $\textbf{b}_j$ and its associated target box $\textbf{b}^t_j$, $N_j$ is the count of foreground pixels in the ground truth mask for $\textbf{b}^t_j$, and $E_{\text{FP}_j}$ \& $E_{\text{FN}_j}$ are proxy measures for the counts of FP and FN pixels in our predicted mask for box $\textbf{b}_j$:
\vspace{-2mm}
\begin{equation}
    E_{\text{FP}_j} = \sum_{xy} \IoU(\mathcal{B}(x,y),\textbf{b}_j) \mathbb{1}_{(x,y)\notin \textbf{M}^t_j},
\end{equation}
\vspace{-4mm}
\begin{equation}
    E_{\text{FN}_j} = \sum_{xy} (1-\IoU(\mathcal{B}(x,y),\textbf{b}_j) )\mathbb{1}_{(x,y)\in \textbf{M}^t_j}.
    \vspace{-2mm}
\end{equation}
$\mathbb{1}_{(x,y)\notin \textbf{M}^t_j}$ and $\mathbb{1}_{(x,y)\in \textbf{M}^t_j}$ are indicator functions representing whether $(x,y)$ belongs to a ground-truth instance mask.
% submission
% where $E_{\text{FP}_j}$ and $E_{\text{FN}_j}$ are proxy measures for the counts of FP and FN pixels in our predicted mask for box $\textbf{b}_j$:
% \begin{equation}
%     E_{\text{FP}_j} = \sum_{xy} \IoU(\mathcal{B}(x,y),\textbf{b}_j) \mathbb{1}_{(x,y)\notin \textbf{M}_{GT_j}}
% \end{equation}
% \begin{equation}
%     E_{\text{FN}_j} = \sum_{xy} (1-\IoU(\mathcal{B}(x,y),\textbf{b}_j) )\mathbb{1}_{(x,y)\in \textbf{M}_{GT_j}},
% \end{equation}
% $\beta_j$ is the IoU between proposal $\textbf{b}_j$ and its associated target box $\textbf{b}^t_j$, and $N_j$ is the count of foreground pixels in the ground truth mask for $\textbf{b}^t_j$.
% ----
%
% We apply the mask loss on the $M$ bounding box proposals after NMS. 
By penalizing FP and FN's, the mask loss helps to improve the final panoptic segmentation result as shown in ablative analysis (Table~\ref{tab:ablative}).
% Our mask loss is similar in spirit to the contrastive loss~\cite{chopra2005learning} for representation learning in that it pulls dense bounding boxes closer to their corresponding query box and pushes them away from query boxes corresponding to other objects.
%\subsection{Loss Function}
% Within the our network, we predict 5 types of information needed for producing final panoptic segmentation results. 

Our final loss function is:
% Accordingly, our final loss function is made of 5 essential components and the optional mask loss as follows. 
\vspace{-2mm}
\begin{equation}
\begin{split}
\label{eq:loss_function}
\mathcal{L}_{total} = \mathcal{L}_{box\_reg} + \mathcal{L}_{center} + \mathcal{L}_{levelness}\\ +  \mathcal{L}_{box\_cls} + \lambda\mathcal{L}_{semantics} + \mathcal{L}_{mask}.
\end{split}
\vspace{-2mm}
\end{equation}
For Cityscapes experiments, we set $\lambda$ to 1 for simplicity. For COCO experiments, we drop $\lambda$ to $0.4$ to account for the increased magnitude of the cross-entropy loss that results from the larger ontology.

\section{Experiments}
\label{sec:experiments}
In this section, we evaluate our method on standard panoptic segmentation benchmarks. We compare our performance to the state of the art in both accuracy and efficiency. We also provide an extensive ablative analysis.

\subsection{Datasets}
% \noindent\textbf{Cityscapes} is comprised of street imagery from Europe and has a total of 5000 densely annotated images with an ontology of 19 classes, 8 thing classes and 11 stuff classes.
The \textbf{Cityscapes} panoptic segmentation benchmark~\cite{cordts2016cityscapes} consists of urban driving scenes with 19 classes, 8 \textit{thing} classes, containing instance level labels, and 11 \textit{stuff} classes.
In our experiments, we only use the images annotated with fine-grained labels: $2975$ for training, $500$ for validation. All the images have a resolution of $1024 \times 2048$.

\textbf{COCO}~\cite{lin2014microsoft} is a large-scale object detection and segmentation dataset. Following the standard protocol, we use the 2017 edition with $118k$ training and $5k$ validation images. The labels consist of 133 classes with $80$ \textit{thing} classes containing instance level annotations.

\subsection{Metrics}
We use the \ac{PQ} metric proposed by~\cite{kirillov2019panoptic} as summary metric: %
\vspace{-1mm}
 \begin{equation}
  PQ = \frac{\Sigma_{(p,g)\in TP}{\IoU}_{(p,g)}\mathbb{1}_{\IoU_{(p,g)}>0.5}}{|TP|+\frac{1}{2}|FP| + \frac{1}{2}|FN|},
  \vspace{-1mm}
\end{equation}
where \textit{p} and \textit{g} are matched predicted and ground-truth segments, and TP, FP, FN denote true positives, false positives, and false negatives, respectively. A positive detection is defined by $\IoU_{(p,g)}>0.5$.
%PQ_dagger
%Considering the limitation of over-penalizes stuff errors in PQ metric as discussed in \cite{porzi2019seamless,li2018learning}, we also report our performance on a augmented metric $PQ^{\dagger}$ that balance the evaluation on stuff, recently proposed in~\cite{porzi2019seamless}.
%
We also provide standard metrics on sub-tasks, including Mean IoU for semantic segmentation and average over $AP^r$~\cite{hariharan2014simultaneous} for instance segmentation.
    % As discussed in \cite{kirillov2018panoptic}, \ac{PQ} can also be decomposed as the product of \ac{SQ} and \ac{RQ},
    % %
    % \begin{equation} \label{eq:PQ-2}
    % \resizebox{.35 \textwidth}{!}{$PQ =\underbrace{ \frac{\Sigma_{(p,g)\in TP}{IoU}_{(p,g)}}{|TP|}}_\text{\acf{SQ}}  +\underbrace{ \frac{|TP|}{|TP|+\frac{1}{2}|FP| + \frac{1}{2}|FN|}}_\text{\acf{RQ}}$}
    % \end{equation}
    
    %To better explore the capability of our proposed approach, we also evaluate our model performance on the task of instance segmentation and semantic segmentation. 
   % For semantic segmentation, we use the standard metric, PASCAL VOC \ac{IoU}.
        % $IoU = TP/ (TP+FP+FN)$
    % $IoUs$ are caculated for each semantic category at pixel level.
    
    %For instance segmentation, following recent literature, we average over the $AP^r$~\cite{hariharan2014simultaneous} with acceptance \ac{IoU} from $0.5$ to $0.95$ in increments of $0.05$. We call this metric $AP$ in the rest of the paper without ambiguity. For minor scores, we also report the $AP50=AP^r(IoU>0.5)$.

%We evaluate our proposed approach using different benchmark datasets, Cityscapes~\cite{cordts2016cityscapes} and COCO~\cite{lin2014microsoft}.
%These datasets have large gaps in both complexity and image content.
% These datasets have similar properties in terms of content, there is a large complexity gap between the two.

% Please add the following required packages to your document preamble:
% \usepackage{graphicx}
\begin{table*}[t]
\centering
% \resizebox{\linewidth}{!}{%
\resizebox{0.90\width}{!}{%
\begin{tabular}{r|c|c|c|c|c|cc|cc}
% \hline
\toprule
Method & Backbone  & PQ & \multicolumn{1}{c|}{$\text{PQ}^{th}$} & $\text{PQ}^{st}$ & \multicolumn{1}{c|}{mIoU} & \multicolumn{1}{c|}{AP} & \multicolumn{1}{c|}{GPU} & \multicolumn{1}{c}{Inference Time}\\ \midrule
\multicolumn{9}{c}{\textbf{Two-Stage}} \\ \midrule
TASCNet~\cite{li2018learning} & ResNet-50-FPN & 55.9 & 50.5 & 59.8 & - & \multicolumn{1}{c|}{-} & V100 & 160ms \\ 
AUNet\cite{li2019attention} & ResNet-50-FPN & 56.4 & 52.7 & 59.0 & 73.6 & \multicolumn{1}{c|}{\underline{33.6}} & - & -\\ 
Panoptic-FPN~\cite{kirillov2019panoptic2} & ResNet-50-FPN   & 57.7 & 51.6& {62.2} & {75.0} & \multicolumn{1}{c|}{{32.0}} & - & - \\ 
$\text{AdaptIS}^{\dagger}$~\cite{sofiiuk2019adaptis} & ResNet-50   & 59.0 & \underline{55.8} & 61.3 & {75.3} & \multicolumn{1}{c|}{32.3} & - & - \\ 
UPSNet~\cite{xiong2019upsnet} & ResNet-50-FPN   & 59.3 & 54.6 & 62.7 & 75.2 & \multicolumn{1}{c|}{33.3} & V100 &$140\text{ms}^*$\\ 
% 
% Seamless Panoptic~\cite{porzi2019seamless} & ResNet-50-FPN   & 59.8  & 54.6& \underline{63.6} & 76.2& \multicolumn{1}{c|}{33.0} & V100 & 150ms \\ 
% % 
Seamless Panoptic~\cite{porzi2019seamless} & ResNet-50-FPN  & \underline{60.2}  & 55.6 & \underline{63.6} &  74.9& \multicolumn{1}{c|}{33.3} & V100 & $150\text{ms}^*$\\ 
 \midrule
% hline for singleshot methods
\multicolumn{9}{c}{\textbf{Single-Stage}} \\ \midrule
DeeperLab~\cite{deeperlab2019} & Wider MNV2   & 52.3 & - & - & - & \multicolumn{1}{c|}{-} & V100 & 251ms  \\ 
FPSNet~\cite{de2019fast} & ResNet-50-FPN   & 55.1 & 48.3 & 60.1 & - & \multicolumn{1}{c|}{-} & TITAN RTX & 114ms \\ 
SSAP~\cite{gao2019ssap} & ResNet-50 & 56.6 & 49.2 & - & - & \multicolumn{1}{c|}{\textbf{31.5}} & 1080Ti & $>$260ms \\ 
DeeperLab~\cite{deeperlab2019} & Xception-71 & 56.5 & - & - & - & \multicolumn{1}{c|}{-} & V100 & 312ms  \\ 
% 
% Ours-R18 & ResNet-18-FPN   & {55.5} & -  & - & - & \multicolumn{1}{c|}{-} & V100 & 77ms \\ 
% %
% Ours-R34 & ResNet-34-FPN   & {56.8} & -  & - & - & \multicolumn{1}{c|}{-} & V100 & 83ms \\ 
% source:a7n1yk62
Ours & ResNet-50-FPN   & \textbf{58.8} &\textbf{52.1}  & \underline{\textbf{63.7}}& \underline{\textbf{77.0}} & \multicolumn{1}{c|}{{29.8}} & V100 & \underline{\textbf{99}}ms \\ 
% \hline
\bottomrule
\end{tabular}%
}
\caption{\textbf{Performance on Cityscapes validation set}. We bold the best number across single-stage methods and underline the best number across the two categories. $\dagger$: method includes multiple-forward passes. $*$: Our replicated result from official sources using the same evaluation environment as our model.}
\label{tab:pq-cityscapes}
\vspace{-3mm}
\end{table*}

% Let's add PQ+ to the paper here:

\subsection{Implementation details}
% Basic platform and hardware
All the models in our experiments are implemented in PyTorch and trained using 8 Tesla V100 GPUs. Inference timing is done on Tesla V100 GPU with batch size 1.

% training schedule
For Cityscapes experiments, our models are trained using a batch size of 1 per GPU, weight decay of $1e^{-4}$, learning rate of $0.013$ for $48$k total steps, decreasing by a factor of 0.1 at step $36$k and $44$k. We apply a random crop of $1800\times900$ and re-scale the crops randomly between $(0.7,1.3)$. 
For COCO experiments, our models are trained using a batch size of 2 per GPU, weight decay of $1e^{-4}$, learning rate of $0.01$ for $180$k steps with learning rate with steps at $120$k and $160$k. For data augmentation, we randomly resize the input image to a shortest side length in $(640,720,800)$. No cropping is applied.
% used for COCO as the dataset doesn't suffer from the same lack of viewpoint variation as Cityscapes. 
% Do I discuss the minimum stuff area dropping or not?

All our models use a ResNet-50 Backbone with ImageNet pretrained weights provided by Pytorch~\cite{NEURIPS2019_9015}. We freeze BatchNorm layers in the backbone for simplicity, similar to~\cite{xiong2019upsnet, li2018learning}.

%-----------------Compare to SOTA Panoptic in turns of accuracy
\subsection{Comparison to State of the Art}
% \input{tables/performance-cityscapes-val.tex}
% \input{tables/coco_performance_comparison.tex}
% Please add the following required packages to your document preamble:
% \usepackage{graphicx}
\begin{table}[t]
\centering
% \resizebox{\linewidth}{!}{%
\resizebox{1\linewidth}{!}{%
\begin{tabular}{r|c|c|c|c|c}
% \hline
\toprule
Method & Backbone & PQ & \multicolumn{1}{c|}{$\text{PQ}^{th}$} & $\text{PQ}^{st}$  & \multicolumn{1}{c}{Inf. Time}\\ \midrule
\multicolumn{6}{c}{\textbf{Two-Stage}} \\ \midrule
Panoptic-FPN~\cite{kirillov2019panoptic2} & ResNet-50-FPN & 33.3  & 45.9 & 28.7 & -  \\ 
$\text{AdaptIS}^{\dagger}$~\cite{sofiiuk2019adaptis} & ResNet-50 & 35.9  & 40.3 & 29.3 & -  \\
AUNet~\cite{li2019attention} & ResNet-50-FPN & 39.6  & \underline{49.1} & 25.2 & -  \\ 
UPSNet~\cite{xiong2019upsnet} & ResNet-50-FPN & \underline{42.5}  & 48.5 & \underline{33.4} & $110\text{ms}^*$   \\ \midrule
\multicolumn{6}{c}{\textbf{Single-Stage}}\\\midrule
DeeperLab~\cite{deeperlab2019} & Xcep-71 & 33.8 & - & -   & 94ms \\ 
SSAP~\cite{gao2019ssap} &  ResNet-50  & 36.5 & - & - & - \\
% source: rduaa33s
Ours & ResNet-50-FPN & \textbf{37.1} & \textbf{41.0} & \textbf{31.3}  & \underline{\textbf{63}}ms \\
\bottomrule
% \hline
\end{tabular}%
}
\caption{\textbf{Performance on COCO-validation}.  We bold the best single-stage methods and underline the best across the two categories.$\dagger$: methods including multiple-forward passes. $*$: Our replicated result from official sources using the same evaluation environment as our model.}
\label{tab:pq-coco}
\vspace{-3mm}
\end{table}
We compare the accuracy and efficiency of our proposed model to the state-of-the-art two-stage and single-stage panoptic segmentation algorithms that use a ResNet-50 or lighter backbone. We only report and compare to single-model prediction results to avoid ambiguity of the inference time during test-time augmentation.
For inference time, we report the average inference time plus NMS processing time over the whole validation set. For Cityscapes, our model takes the full resolution as input size. For COCO, we resize all images to a longer side of $1333$px as input. 
Our quantitative results are reported in Table~\ref{tab:pq-cityscapes} and Table~\ref{tab:pq-coco}.

% highlight our advances
Our model outperforms all the single-stage methods by a significant margin in both accuracy and inference speed. We are also closing the gap between the single-stage and slow but state-of-the-art two-stage methods, even outperforming some of them.
To better highlight the potential of our method towards deployment in real-world systems and its parallelization benefits, we conduct a simple optimization by compiling the convolutional layers in our model using TensorRT~\cite{tensorrt}. It enables our model to operate in real-time ($30.3$ FPS on a V100 GPU) at full resolution on Cityscapes videos vs.~$10.1$ FPS without optimization (cf.~Figure~\ref{fig:infer_time}).

\begin{figure*}[bp]
\begin{center}
\def\figwidth{0.29}
% \includegraphics[width=\figwidth\linewidth]{figures/source/rgb/frankfurt_000001_032711_leftImg8bit.png}
% \includegraphics[width=\figwidth\linewidth]{figures/source/panoptic/frankfurt_000001_032711.png}
% \includegraphics[width=\figwidth\linewidth]{figures/source/upsnet/frankfurt_000001_032711.png}

% bad example
% \includegraphics[width=\figwidth\linewidth]{figures/source/rgb/frankfurt_000001_031266_leftImg8bit.png}
% \includegraphics[width=\figwidth\linewidth]{figures/source/panoptic/frankfurt_000001_031266.png}
% \includegraphics[width=\figwidth\linewidth]{figures/source/upsnet/frankfurt_000001_031266.png}

%
\includegraphics[width=\figwidth\linewidth]{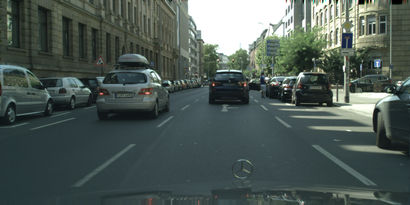}
\includegraphics[width=\figwidth\linewidth]{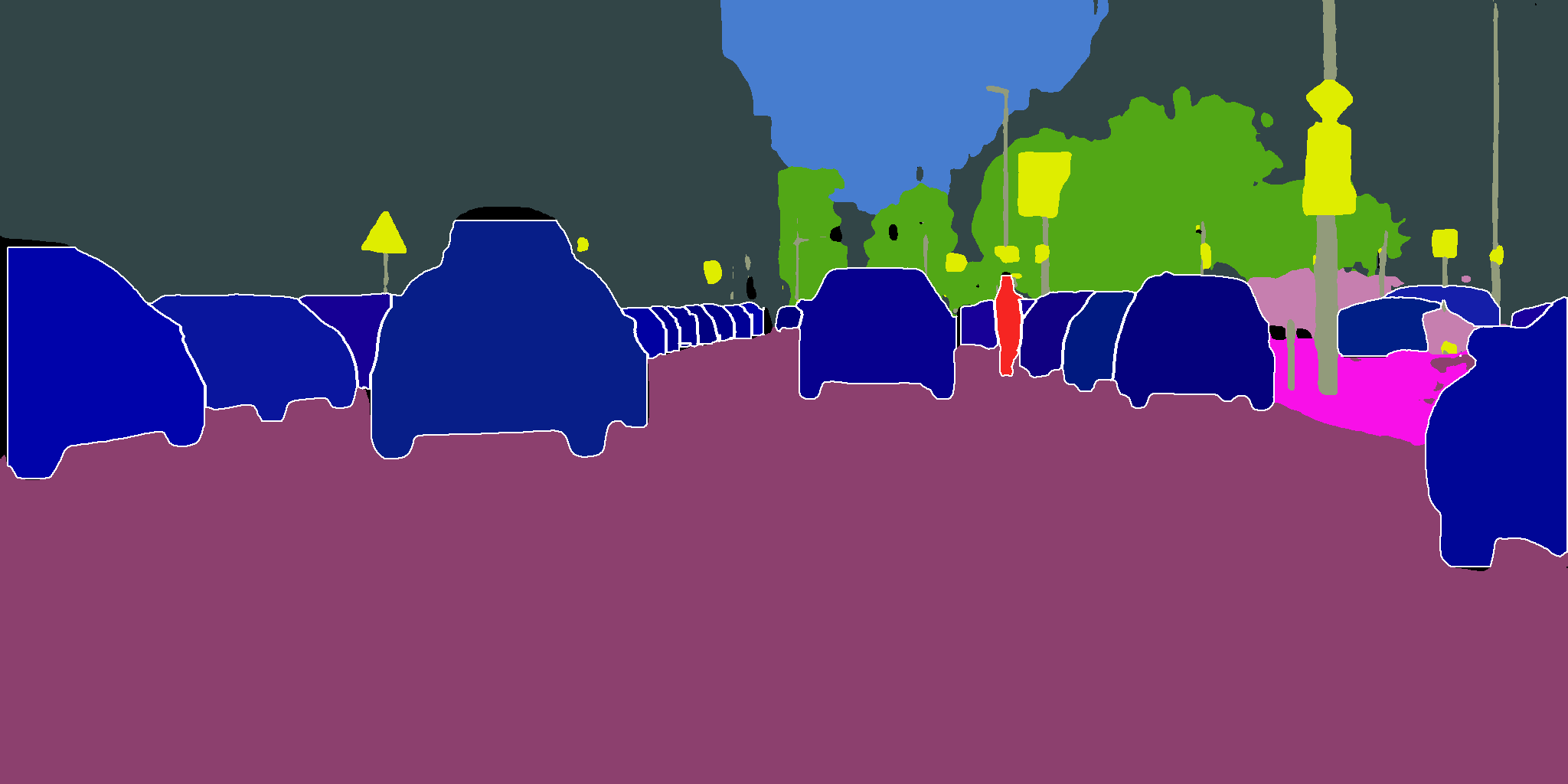}
\includegraphics[width=\figwidth\linewidth]{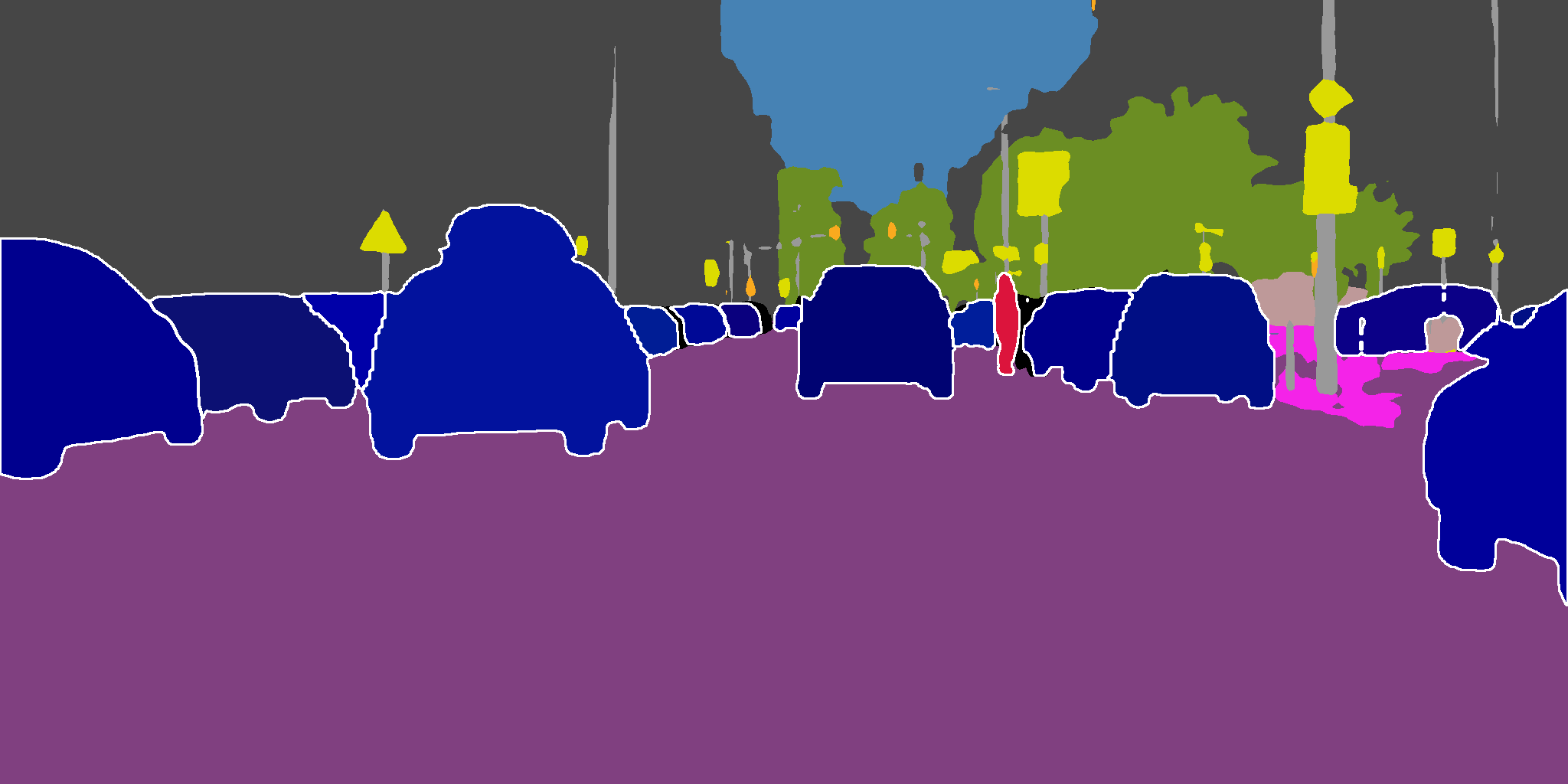}

\includegraphics[width=\figwidth\linewidth]{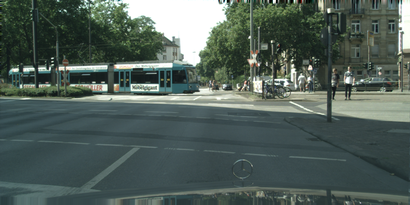}
\includegraphics[width=\figwidth\linewidth]{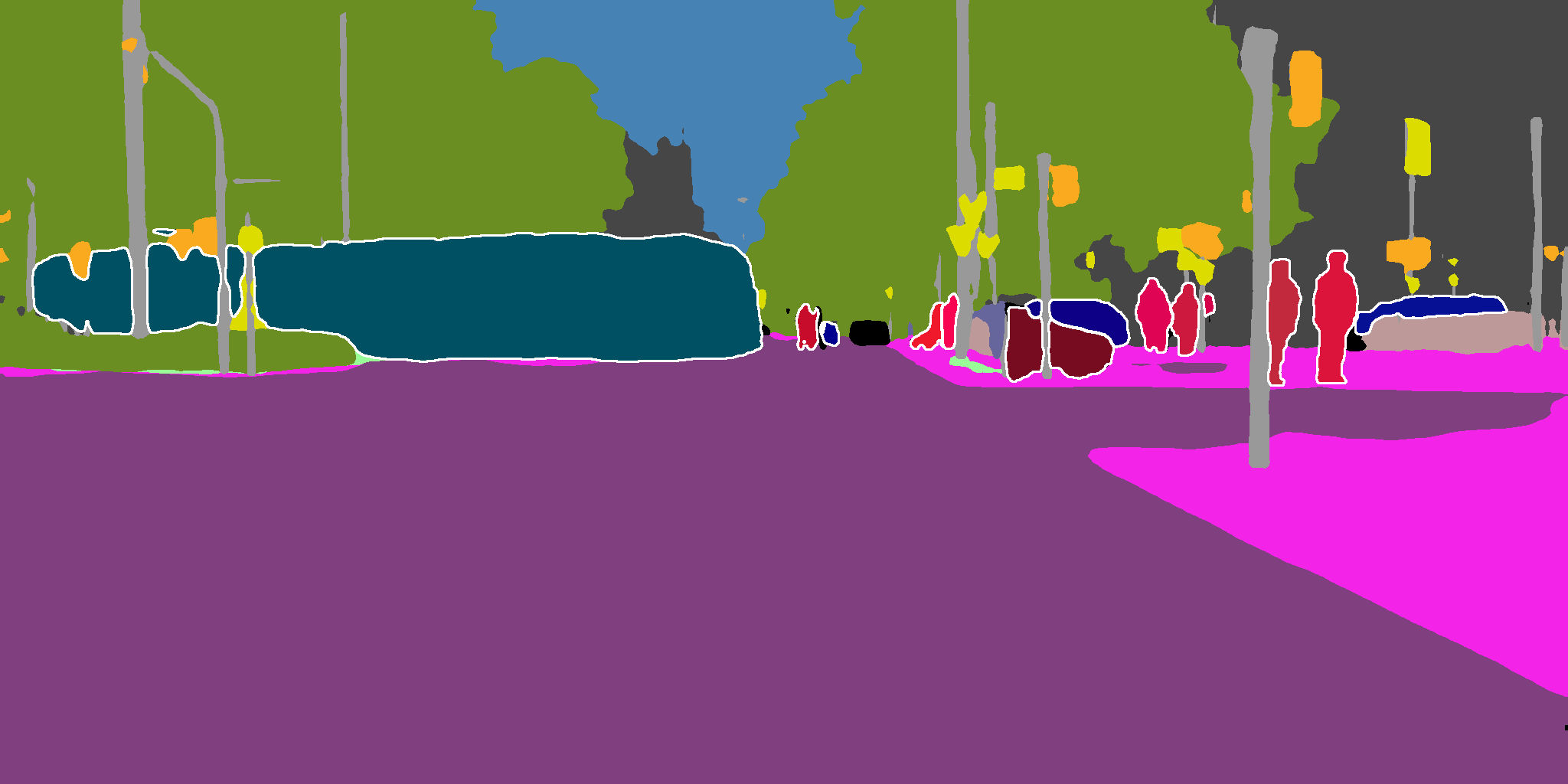}
\includegraphics[width=\figwidth\linewidth]{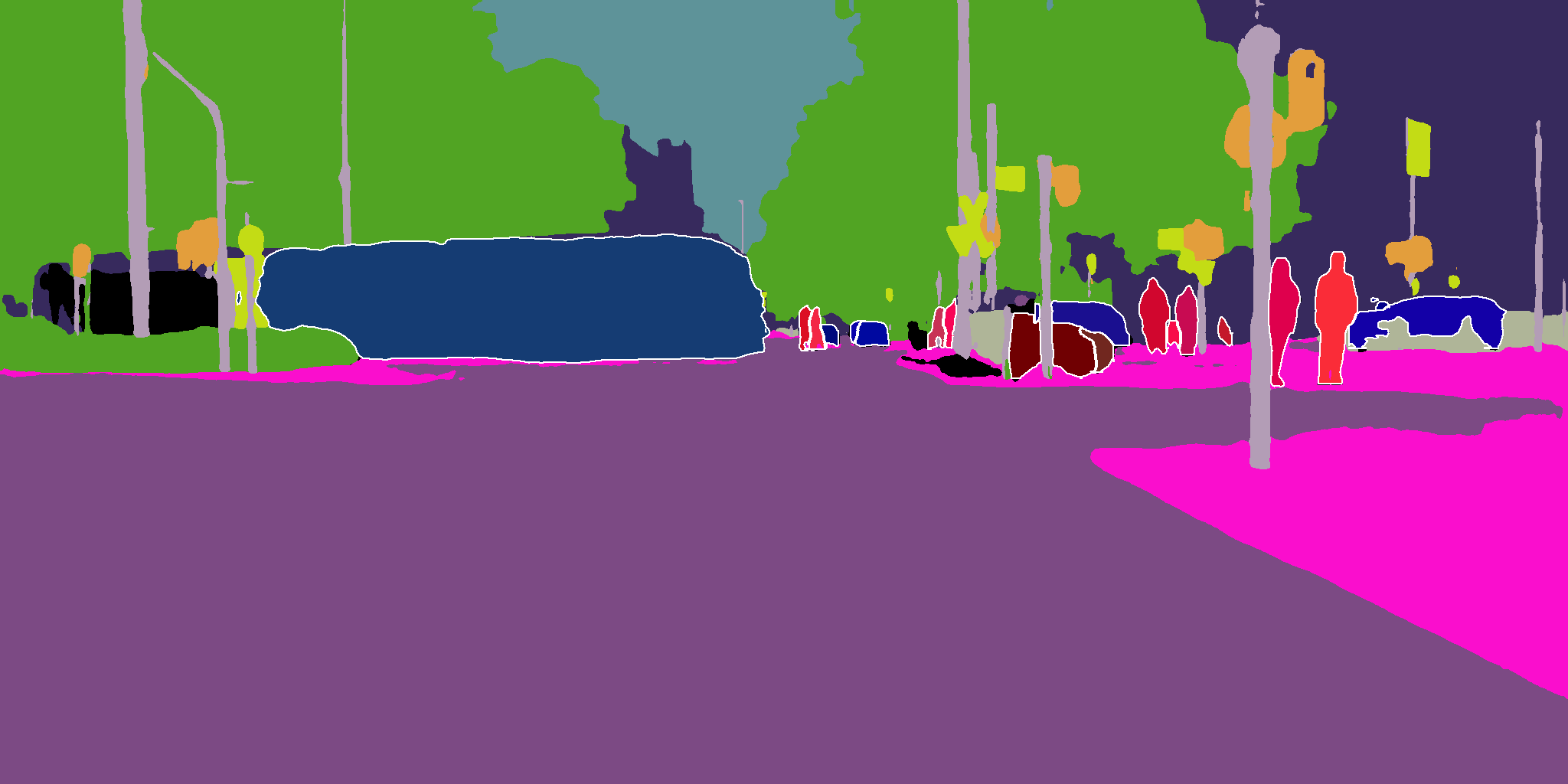}
\includegraphics[width=\figwidth\linewidth]{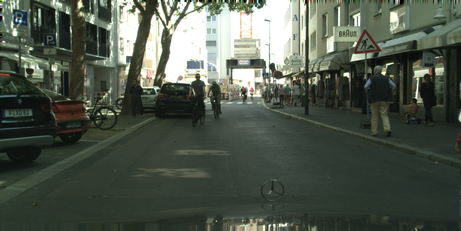}
\includegraphics[width=\figwidth\linewidth]{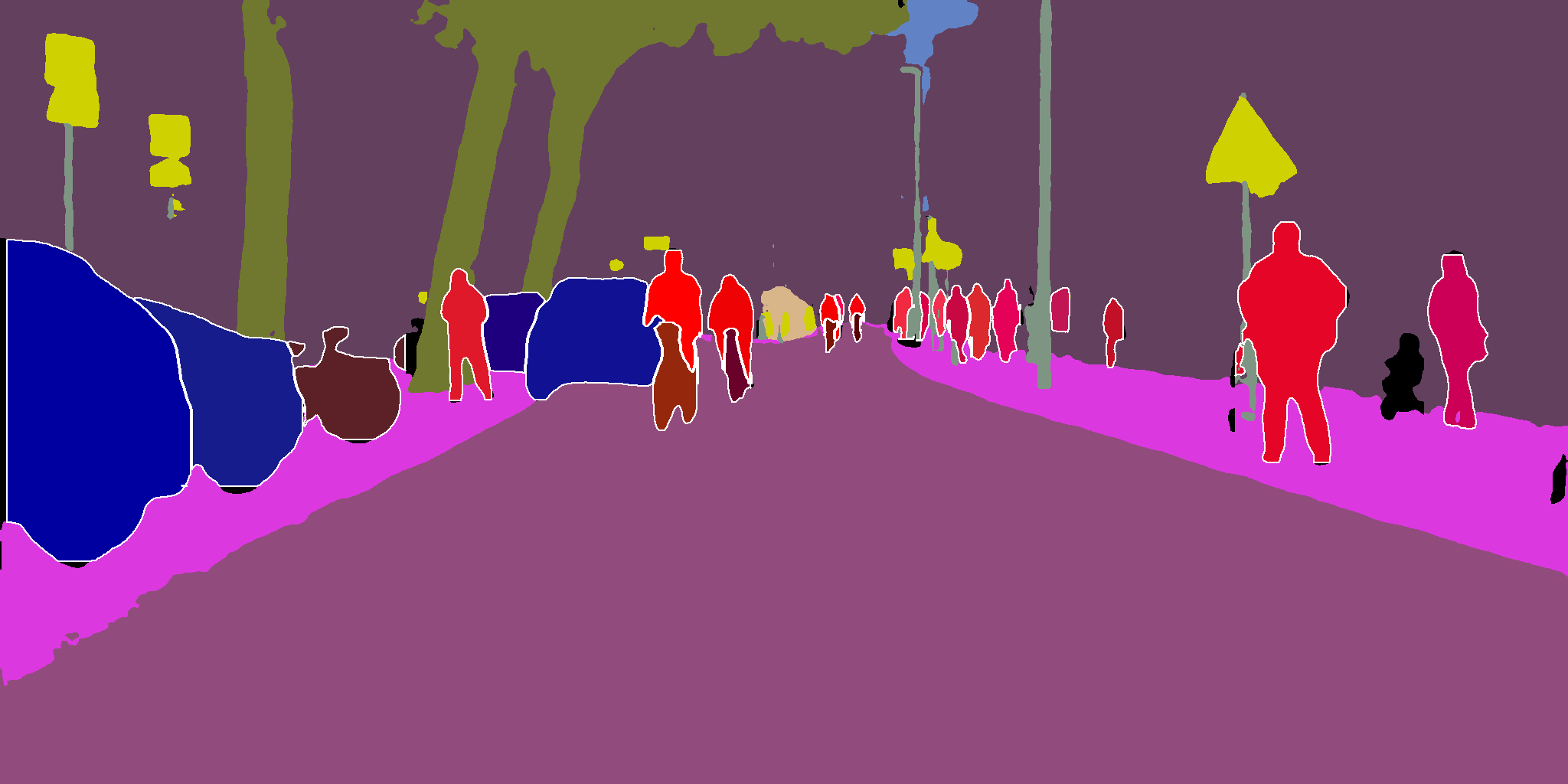}
\includegraphics[width=\figwidth\linewidth]{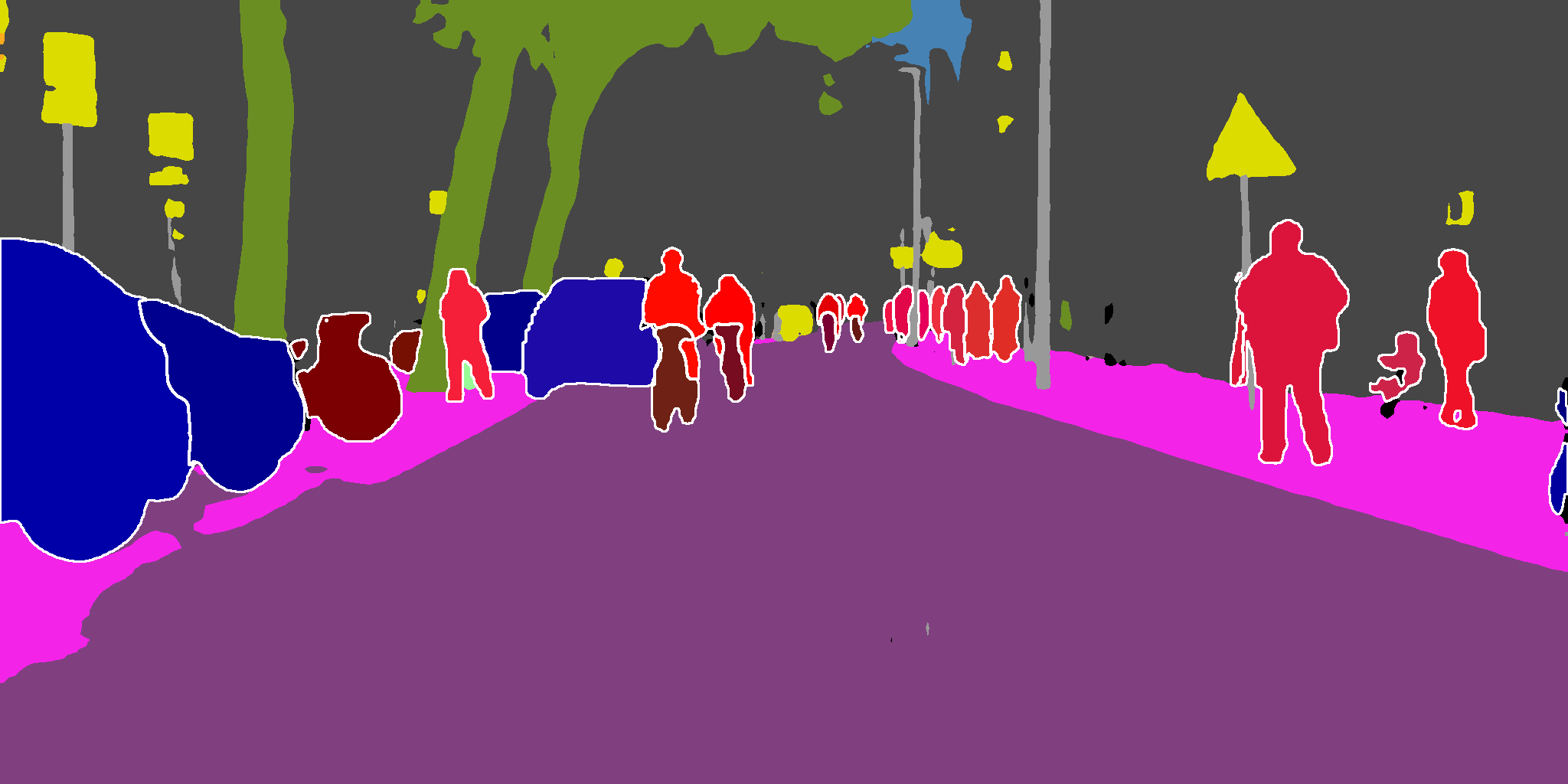}

% % %
% \includegraphics[width=\figwidth\linewidth]{figures/source/rgb/frankfurt_000001_002646_leftImg8bit.png}
% \includegraphics[width=\figwidth\linewidth]{figures/source/panoptic/frankfurt_000001_002646.png}
% \includegraphics[width=\figwidth\linewidth]{figures/source/upsnet/frankfurt_000001_002646.png}

%
\includegraphics[width=\figwidth\linewidth]{figures/source/rgb/000000006954.jpg}
\includegraphics[width=\figwidth\linewidth]{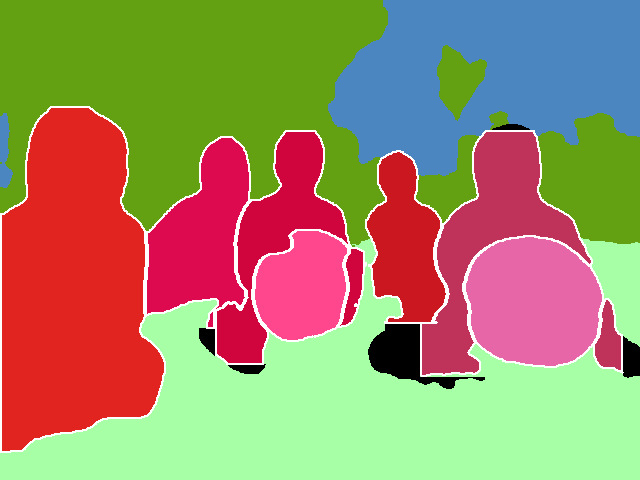}
\includegraphics[width=\figwidth\linewidth]{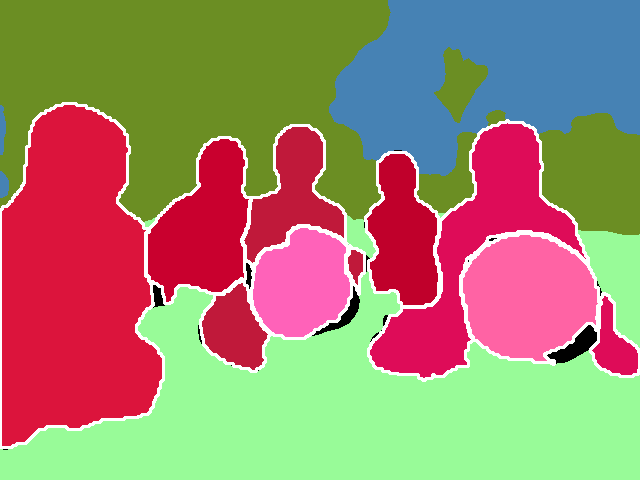}

\includegraphics[width=\figwidth\linewidth]{figures/source/rgb/000000216296.jpg}
\includegraphics[width=\figwidth\linewidth]{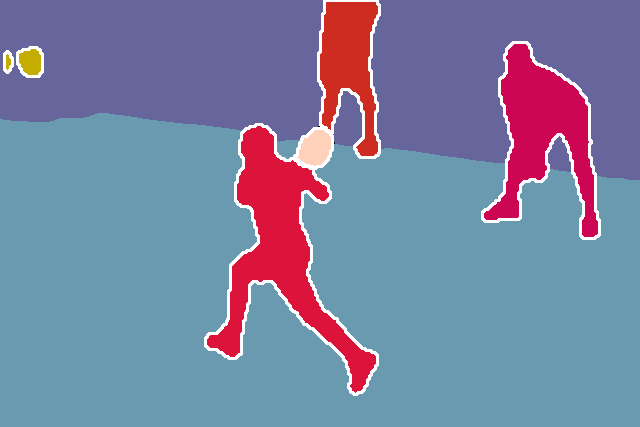}
\includegraphics[width=\figwidth\linewidth]{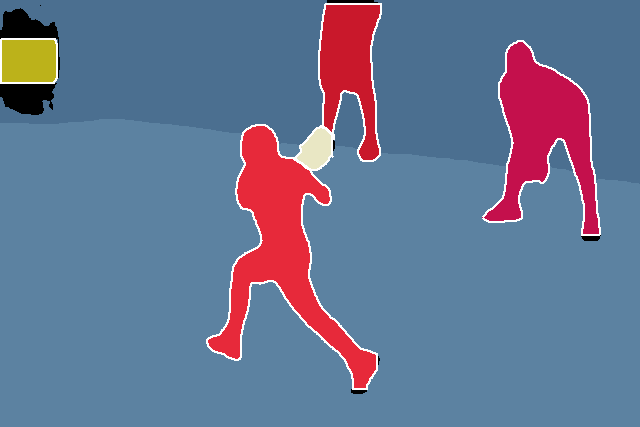}
\end{center}
\vspace{-4mm}
   \caption{\textbf{Panoptic segmentation results} on CityScapes and COCO comparing our predictions and UPSNet~\cite{xiong2019upsnet}. We leave it to the reader to guess which results are ours (the answer is in \protect\cite{answer}).}
\label{fig:cityscales-val-visualization}
\end{figure*}

We also provide some qualitative examples comparing to one of the best two-stage methods, UPSNet~\cite{xiong2019upsnet}, in Figure~\ref{fig:cityscales-val-visualization}. Our method presents little degradation compared to UPSNet. In fact, our models provide better mask estimation on rare-shape instances (cf.~the car with a roof-box, or the small child) thanks to the self-attention in mask association. We observe particularly good results on unusual shapes, because unlike Mask-RCNN style methods\cite{he2017mask}, our mask construction process can associate pixels with a detection even if they fall outside its bounding box.

% inference time discussion
% To better evaluate our model performance with regarding to inference time, we also provide more thorough comparison to most of the single-stage methods and some open source two-stage methods in Table~\ref{tab:inference-cityscapes} and Table~\ref{tab:inference-coco} under different configurations. The experimental results further indicate that our model provides a better trade-off curve between inference time and panoptic segmentation accuracy.

%% -----------------Inference time analysis, comparison and discussion
% \subsection{Inference time analysis}
% inference time comparion on pytorch GPU
% \input{tables/inference-time-table-pytorch.tex}
% \input{tables/inference-time-table-python-coco.tex}

\subsection{Ablative Analysis}
\vspace{-5pt}
\label{sec:ablative}
% \begin{table}[]
% \centering
% \resizebox{\linewidth}{!}{%
% \begin{tabular}{c|c|c|c|ccc}
% \hline
% split tower & Levelness & sem. prob. & mask loss & \multicolumn{1}{c|}{$\text{PQ}$} & \multicolumn{1}{c|}{$\text{PQ}_{\text{th}}$} & $\text{PQ}_{\text{st}}$ \\ \hline
% \multicolumn{7}{c}{\textbf{Fully Supervised}} \\ \hline
% % source: 6wqekdq3
% \hline
%  &  &  &  & 56.8 & 48.1 & 63.6 \\ \hline
%  % source: hy1g53wh
%  \checkmark&  &  &  & 57.1 &  47.8 & \textbf{63.8} \\ \hline
%  % source: 
%  \checkmark& \checkmark &  &  & 58.0 & 50.0 & 63.7 \\ \hline
%  % source: t0u75kct
%  \checkmark& \checkmark &  \checkmark&  & 58.1 & 50.4 & 63.7 \\ \hline
%  % source:a7n1yk62
%  \checkmark& \checkmark &  \checkmark&  \checkmark& \textbf{58.8} & \textbf{52.1} &{63.7}  \\ \hline
% \multicolumn{7}{c}{\textbf{Weakly Supervised (No mask label)}} \\ \hline
% % source:ekym9oki
%  \checkmark& \checkmark &  \checkmark&  & 55.7 & 45.2 & 63.3 \\ \hline
% \end{tabular}%
% }
% \caption{\textbf{Ablative analysis} We compare the impact of different key modules/designs in our proposed network. We also present weakly supervised model trained with only semantic segmentation and bounding box labels.}
% \label{tab:ablative}
% \vspace{-5mm}
% \end{table}
\begin{table}[h]
\centering
\resizebox{\linewidth}{!}{%
\begin{tabular}{c|c|c|c|c|c}
% \hline
\toprule
Two towers & Levelness & Mask loss & \multicolumn{1}{c|}{$\text{PQ}$} & \multicolumn{1}{c|}{$\text{PQ}^{th}$} & $\text{PQ}^{st}$ \\ \hline
\multicolumn{6}{c}{\textbf{Fully Supervised}} \\ \hline
% source: 6wqekdq3

 &  &  & 56.8 & 48.1 & 63.1 \\ \hline
 % source: hy1g53wh
 \checkmark&  &  & 57.1 &  47.8 & \textbf{63.8} \\ \hline
 % source: t0u75kct
 \checkmark& \checkmark &  &  58.1 & 50.4 & 63.7 \\ \hline
 % source: t0u75kct
 %  \checkmark& \checkmark &  & 58.1 & 50.4 & 63.7 \\ \hline
 % source:a7n1yk62
 \checkmark& \checkmark &  \checkmark& \textbf{58.8} & \textbf{52.1} &{63.7}  \\ \hline
\multicolumn{6}{c}{\textbf{Weakly Supervised (No mask label)}} \\ \hline
% source:ekym9oki
 \checkmark& \checkmark &  & 55.7 & 45.2 & 63.3 \\ 
 %\hline
 \bottomrule
\end{tabular}%
}
\caption{\textbf{Ablative analysis}. We compare the impact of different key modules/designs in our proposed network. We also present a weakly supervised model trained without using instance masks.}
\label{tab:ablative}
\vspace{-5mm}
\end{table}
We provide an ablative analysis on the key modules of our method in Table~\ref{tab:ablative}.

The first row presents a simplified baseline with \textit{one} feature tower in the panoptic head that is used for both localization and semantics. We note that this baseline already leads to strong performance that is better than some two-stage methods reported in Table~\ref{tab:pq-cityscapes}. This architecture can be used to achieve even greater speedups. In the second row, we can see how using two separate feature towers, one for localization and one for semantics, improves model performance slightly. 

Then, we introduce levelness, which leads to almost a $1$ point boost in $\text{PQ}$.
Without levelness, it is still possible to compute $\hat{P}_{loc}$ as in Eq.~\ref{eq:ploc}. We can compute the $IoU$ between each query box and the predicted bounding boxes from every FPN level, and then take a max along the FPN levels. However, this operation suffers from ambiguity between object boundaries and background.

Finally, in the fourth row, we introduce our explicit mask loss from Section~\ref{sec:mask_loss} which further refines mask association and construction for \textit{thing} classes resulting in a higher $\text{PQ}^{\text{th}}$ and corresponding bump in $\text{PQ}$.
% We compare different configurations to our proposed method starting from a shared tower in panoptic head.

\subsection{Weakly supervised extension}
We provide a simple yet promising extension of our model to the weakly supervised setting in which we have ground truth bounding boxes and semantic segmentation but no instance masks. 
%Since we can convert panoptic segmentation from semantic segmentation and dense detections, our model can be trained with \textbf{only} semantic labels and bounding box labels. 
A few straightforward changes are required: (1) regarding bounding box target assignment described in section 4.1, we now consider a pixel foreground as long as it is inside a ground truth bounding box, (2) we update the foreground indicator function $\mathbb{1}_{\text{\textbf{fg}}}(x, y)$ in Eq.~\ref{eq:l_box_reg} and Eq.~\ref{eq:l_center} from `in-mask' to `in-box', and (3) we train without using the Mask Loss. Our weakly supervised Cityscapes model achieves promising accuracy and even outperforms some fully supervised  methods, as shown in the last row of Table~\ref{tab:ablative}.
% More explorations can be done along this direction on our proposed method. We leave it to the future work. 
% In Figure~\ref{fig:property}, we give real examples highlighting unique advances in our algorithms. As shown in the examples, our panoptic transforming process is able to recovered from imperfect bounding box proposals as well as handling multiple contours due to occlusion.

%Please refer to additional configuration, visualization and detail metrics of our approach in supplemental material.
% \input{figures/property-example.tex}
%\subsection{Weakly supervised application}\label{sec:weakly-supervised}

% \footnotetext{We put ours in the middle for row 2,4,6 and on the right 1,3,5.}
% \footnotetext{We put ours in the middle for row 1,3,5 and on the right 2,4,6.}

\section{Conclusion}
In this work, we propose a single-stage panoptic segmentation framework that achieves real-time inference with a performance competitive with the current state of the art. 
We first introduce a novel parameter-free mask construction operation that reuses predictions from dense object detection via a global self-attention mechanism. Our architecture dramatically decreases computational complexity associated with instance segmentation in conventional panoptic segmentation algorithms. 
Additionally, we develop an explicit mask loss which improves panoptic segmentation quality.
Finally, we evaluate the potential of our method in weakly supervised settings, showing that it can outperform some recent fully supervised methods.
%
% Our qualitative results suggest that further improvements are mostly expected on thin boundaries and small objects. Increasing precision while retaining real-time inference in those fine-grained cases is still an open problem.

\section*{Acknowledgments}
% This research was supported by the Toyota Research Institute (TRI). This article solely reflects the opinions and conclusions of its authors and not TRI or any other Toyota entity. 
We would like to thank the TRI ML team, especially Dennis Park and Wolfram Burgard for their support and insightful comments during the development of this work. 

\clearpage
\balance
\renewcommand\tablename{Table~A}
\renewcommand\figurename{Figure~A}
% Remove white space between "S" and number:
\renewcommand\thetable{\unskip\arabic{table}}
\renewcommand\thefigure{\unskip\arabic{figure}}
\def\theequation{A\arabic{equation}}
\section*{Appendix}
\appendix
We provide more detailed descriptions, metrics, and visualizations for our proposed approach that were not included in the main text due to space limitation:
\vspace{-3mm}
\begin{itemize}
    \item per-class metrics of our model on Cityscapes and COCO (Tables 1 and 2 in the main text);
    \vspace{-2mm}
    \item implementation details for the comparison without levelness presented in the ablative analysis (Table 3 in the main text);
    \vspace{-2mm}
    \item detailed description and discussion of the proposed weakly supervised application of our method (Section 5.6 Table 3 in the main text);
    \vspace{-2mm}
    \item discussion on the degradation curve of our method with lighter backbones.
\end{itemize}

\section{Per-class Performance}

We provide per-class PQ metrics of our models on Cityscapes and COCO dataset in \tablename\ref{tab:cityscapes-detail} and \tablename\ref{tab:coco-detail}, which corresponds respectively to the entries in Table 1 and Table 2 in the main text.
For both datasets, $\sigma=0.3$ is used as the foreground mask acceptance probability in Eq. 8, which is determined through hyper-parameter grid search.
\section{Contribution of Levelness Map}

As we have shown in the paper (also depicted in Figure~3), our proposed model predicts a global levelness map using feature maps produced by the localization tower at all scale levels.
This levelness map indicates which scale level does the bounding box at each location $(x,y)$ belongs to (with 0 reserved for background). 
At inference time, the levelness map is used to provide indexes while we assemble the global dense bounding box prediction $\mathcal{B}(x,y)$ from each FPN level ($\{\textbf{b}^i_{xy}\}$), according to Eq.~15.

However, the levelness is not a necessity in our proposed model, as the assembling process can be done without it. In our ablative analysis in Section 5.5, we compared our method to a simple alternative assembling approach without levelness to justify the value of such design.  

This alternative solution, instead of assembling a unique bounding box for each feature location $(x,y)$, carries all the bounding box predictions from different FPN levels $i$ to the foreground mask probability estimation:
\begin{equation}
    \mathcal{B}(x,y;i) = \textbf{b}^i_{xy}
\end{equation}{}
Then the location based foreground probability becomes:
\begin{equation}
    \hat{P}_{loc}(x,y,j;i) = \IoU( \mathcal{B}(x,y;i),\textbf{B}_j)
\end{equation}{}
We can still obtain a single probability for each location by taking the maximum along the level dimension $i$:
\begin{equation}
    \hat{P}_{loc}(x,y,j) = \argmax_i( \hat{P}_{loc}(x,y,j))
\end{equation}{}
% This levelness map indicates which scale level each location $(x,y)$ belongs to (with 0 reserved for background). 
% With this levelness map, at inference time, we first upscale dense bounding box predictions at all scale levels to the same size of the levelness map, namely $(H/4, W/4)$.

% Next, according to the classification results obtained in the levelness map, we assemble the final dense bounding box map from the up-scaled multi-scale bounding box predictions for instance mask recovery.

% We can also avoid the prediction and use of a levelness map. 
% Instead of a flat map of dense bounding box predictions with a size of $(H/4, W/4)$, we can directly use a tensor of dense bounding boxes of size $(H/4, W/4, 5)$, where 5 is the number of scale levels in our setting. 
% At inference, to recover instance masks with query bounding boxes and dense bounding boxes, for each query bounding box $\textbf{B}_j$, at each location $(x, y)$, we first extract the best bounding box (i.e.\ with the maximum IoU with $\textbf{B}_j$) along the level dimension  and then recover instance masks following the same procedure as using a levelness map.
We report the resulting model performance of this alternative solution in the second entry of the ablative analysis table (Table 3 in the main text).
% To investigate the effect of levelness, we train two versions of models with the only difference being the use of levelness.
For simplicity, this comparison on levelness is done without the mask loss.
All other configurations are the same as the default model.
This comparison indicates that the levelness map, with only 1 additional convolutional layer, is able to provide a better cross-level indication resulting in an increase in performance (+1$\%$ PQ). 
% \begin{table}[]
% \centering
% \resizebox{0.3\textwidth}{!}{%
% \begin{tabular}{@{}rccc@{}}
% \toprule
% \textbf{Class} & \textbf{PQ} & \textbf{SQ} & \textbf{RQ} \\ \midrule
% Mean & 58.8 & 79.8   &72.3  \\
% \hline
% road & 97.7 & 98.0 & 99.7 \\
% sidewalk & 75.3 & 83.8 & 89.9 \\
% building & 87.8 & 89.4 & 98.2 \\
% wall & 25.5 & 71.2 & 35.8 \\
% fence & 27.0 & 71.5 & 37.7 \\
% pole & 48.3 & 64.6 & 74.8 \\
% traffic-light & 44.7 & 68.6 & 65.2 \\
% traffic-sign & 66.1 & 75.9 & 87.0 \\
% vegetation & 88.6 & 90.3 & 98.2 \\
% terrain & 28.9 & 73.4 & 39.3 \\
% sky & 86.2 & 92.3 & 93.4 \\
% person & 51.6 & 76.9 & 67.0 \\
% rider & 53.5 & 71.7 & 74.6 \\
% car & 64.8 & 83.2 & 77.8 \\
% truck & 53.3 & 86.3 & 61.7 \\
% bus & 72.9 & 88.5 & 82.4 \\
% train & 65.6 & 83.1 & 78.9 \\
% motorcycle & 44.5 & 72.3 & 61.5 \\
% bicycle & 42.2 & 71.6 & 58.9 \\ \bottomrule
% \end{tabular}
% }
% \end{table}

\begin{table}[ht]
\centering
\resizebox{0.45\textwidth}{!}{%
\begin{tabular}{@{}rccc|rccc@{}}
\toprule
{Class} & {PQ} & {SQ} & {RQ}&{Class} & {PQ} & {SQ} & {RQ} \\ \midrule
\textbf{Mean} & 58.81 & 79.81   &72.32 &road & 97.58 & 97.88 & 99.69\\
sidewalk & 75.91 & 83.87 & 90.51 & building & 87.67 & 89.49 & 97.96\\
wall & 30.02 & 72.21 & 41.57 & fence & 34.89 & 73.27 & 47.62\\
pole & 50.15 & 65.23 & 76.88 & traffic light & 45.98 & 70.97 & 64.79\\
traffic sign & 68.23 & 77.32 & 88.24 & vegetation & 88.78 & 90.27 & 98.35\\
terrain & 34.45 & 73.75 & 46.72 & sky & 86.73 & 92.16 & 94.10\\
person & 47.96 & 75.76 & 63.30& rider & 49.61 & 70.76 & 70.11\\
car & 60.55 & 83.30 & 72.69 & truck & 47.16 & 83.12 & 56.74\\
bus & 68.51 & 88.75 & 77.19 & train & 55.94 & 83.91 & 66.67\\
motorcycle & 44.54 & 72.68 & 61.28 & bicycle & 42.86 & 71.76 & 59.72\\
\bottomrule
\end{tabular}
}
\caption{\textbf{Per-class Performance on Cityscapes}}
\label{tab:cityscapes-detail}
\end{table}
\section{Weakly Supervised Application}

In the weakly supervised scenario discussed in Section 5.6 of the main text, we consider relying only on semantic and bounding box labels for the panoptic segmentation task. 
% \begin{figure}[h]
%     \centering
%     \begin{subfigure}[]{0.5\linewidth}
%         \includegraphics[width=0.95\linewidth]{figures/source/weak-label-source.png}
%         \label{fig:box-compensation}
%     \end{subfigure}%
%     \begin{subfigure}[]{0.5\linewidth}
%         \includegraphics[width=0.95\linewidth]{figures/source/weak-label.png}
%         \label{fig:occlusion}
%     \end{subfigure}
% \caption{\textbf{Weak Supervision} RGB image (left) and Weak supervision used (right). }
% \label{fig:weak-label}
% \end{figure}

\begin{figure}[h]
    \centering
    \begin{subfigure}[]{0.33\linewidth}
        \includegraphics[width=0.95\linewidth]{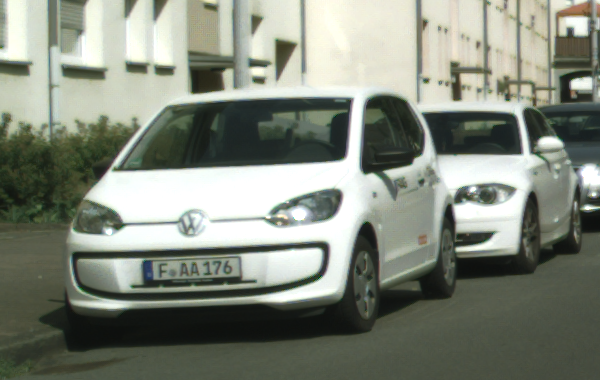}
    \end{subfigure}%
    \begin{subfigure}[]{0.33\linewidth}
        \includegraphics[width=0.95\linewidth]{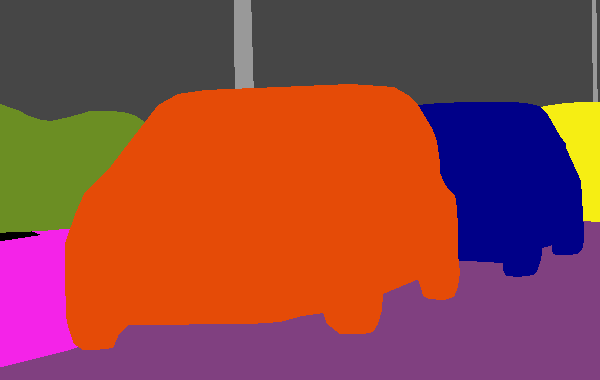}
    \end{subfigure}%
    \begin{subfigure}[]{0.33\linewidth}
        \includegraphics[width=0.95\linewidth]{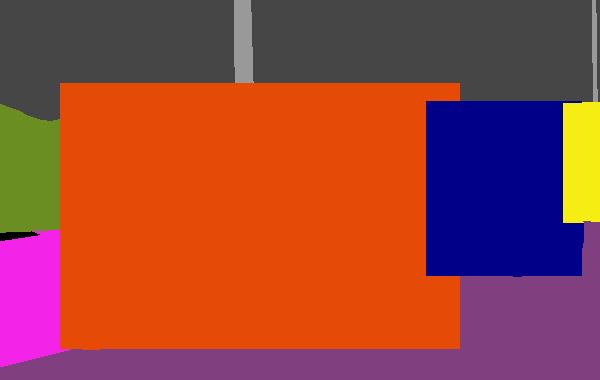}
    \end{subfigure}
\caption{\textbf{Weak Supervision} Left: input image; Middle: fully supervised pixel association; Right: Weakly supervised pixel association. In this example, orange, blue and yellow color indicates the pixels which are assigned to a bounding box target of the three cars during training.}
\label{fig:weak-label}
\end{figure}
\begin{figure}[ht]
\begin{center}
\def\figwidth{0.45}

\includegraphics[width=\figwidth\linewidth]{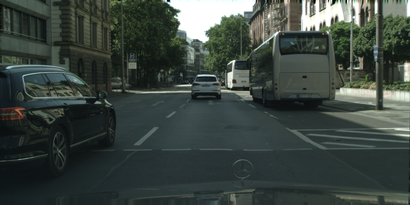}
\includegraphics[width=\figwidth\linewidth]{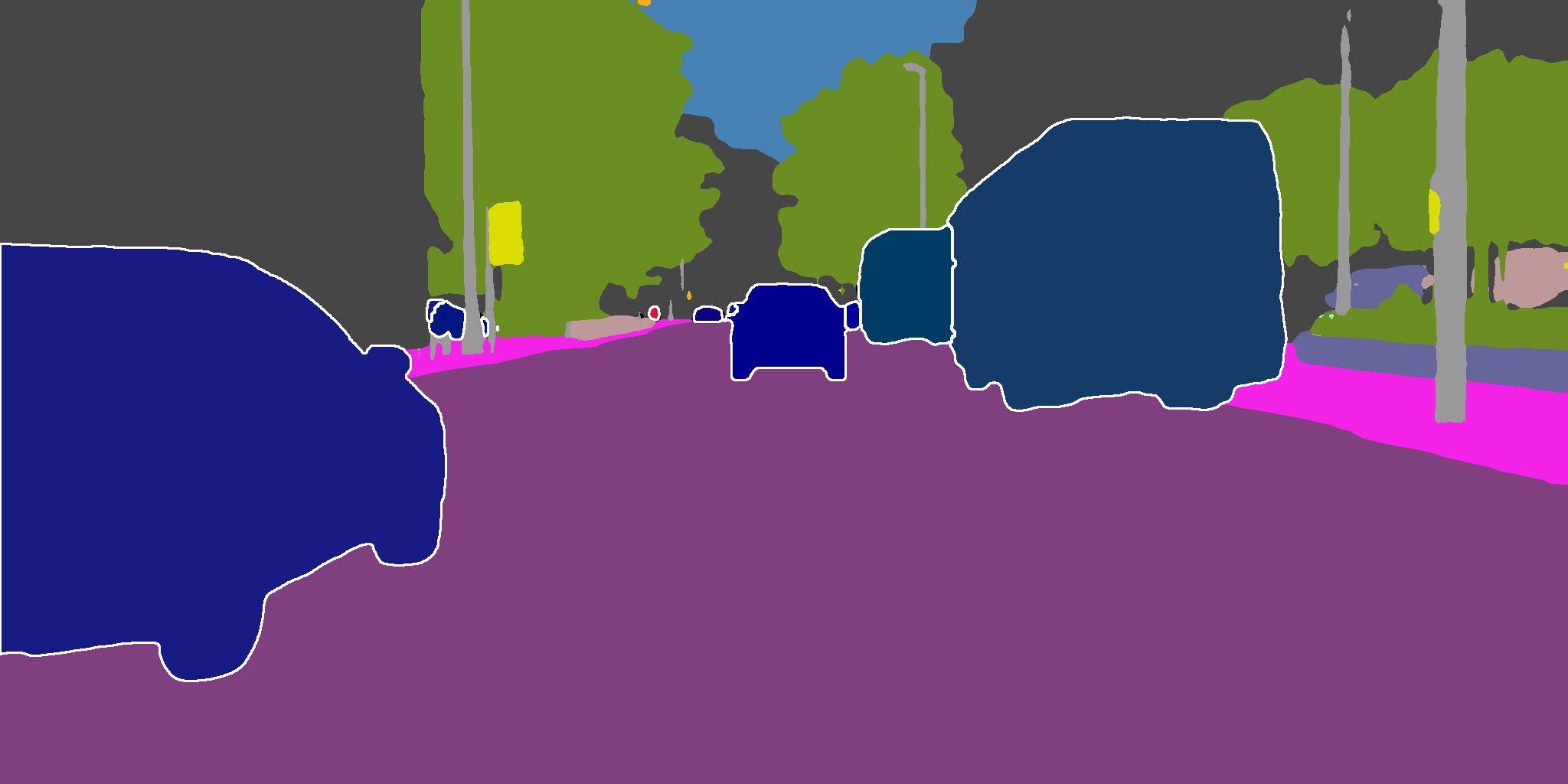}

\includegraphics[width=\figwidth\linewidth]{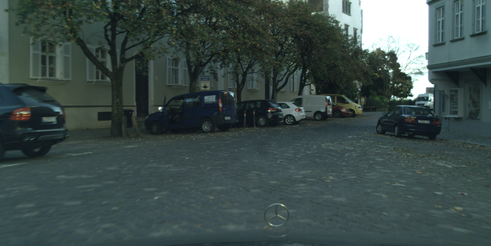}
\includegraphics[width=\figwidth\linewidth]{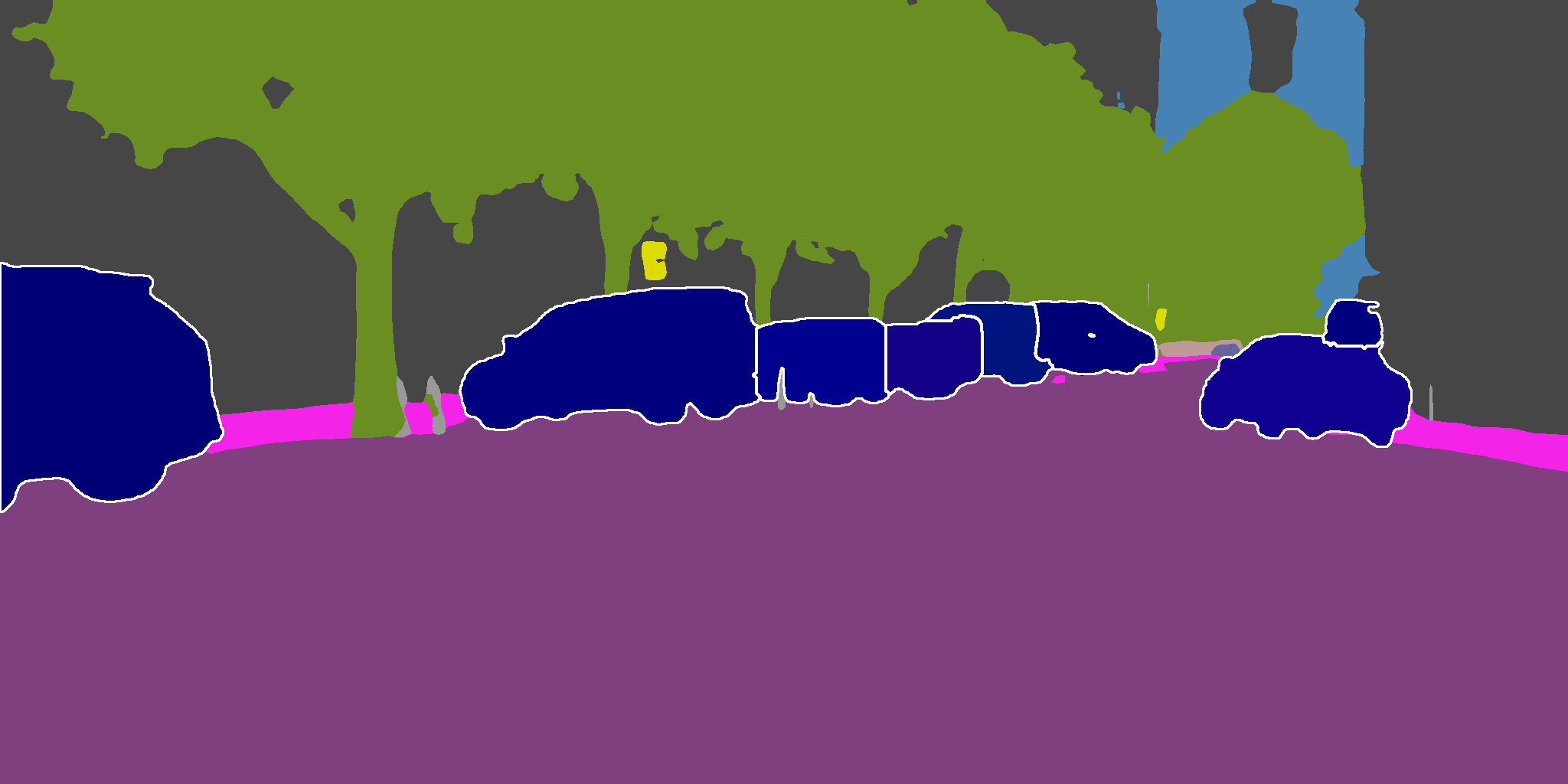}

\includegraphics[width=\figwidth\linewidth]{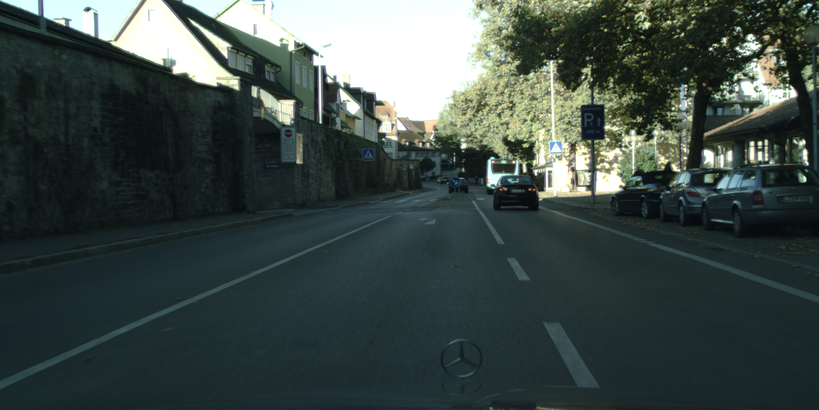}
\includegraphics[width=\figwidth\linewidth]{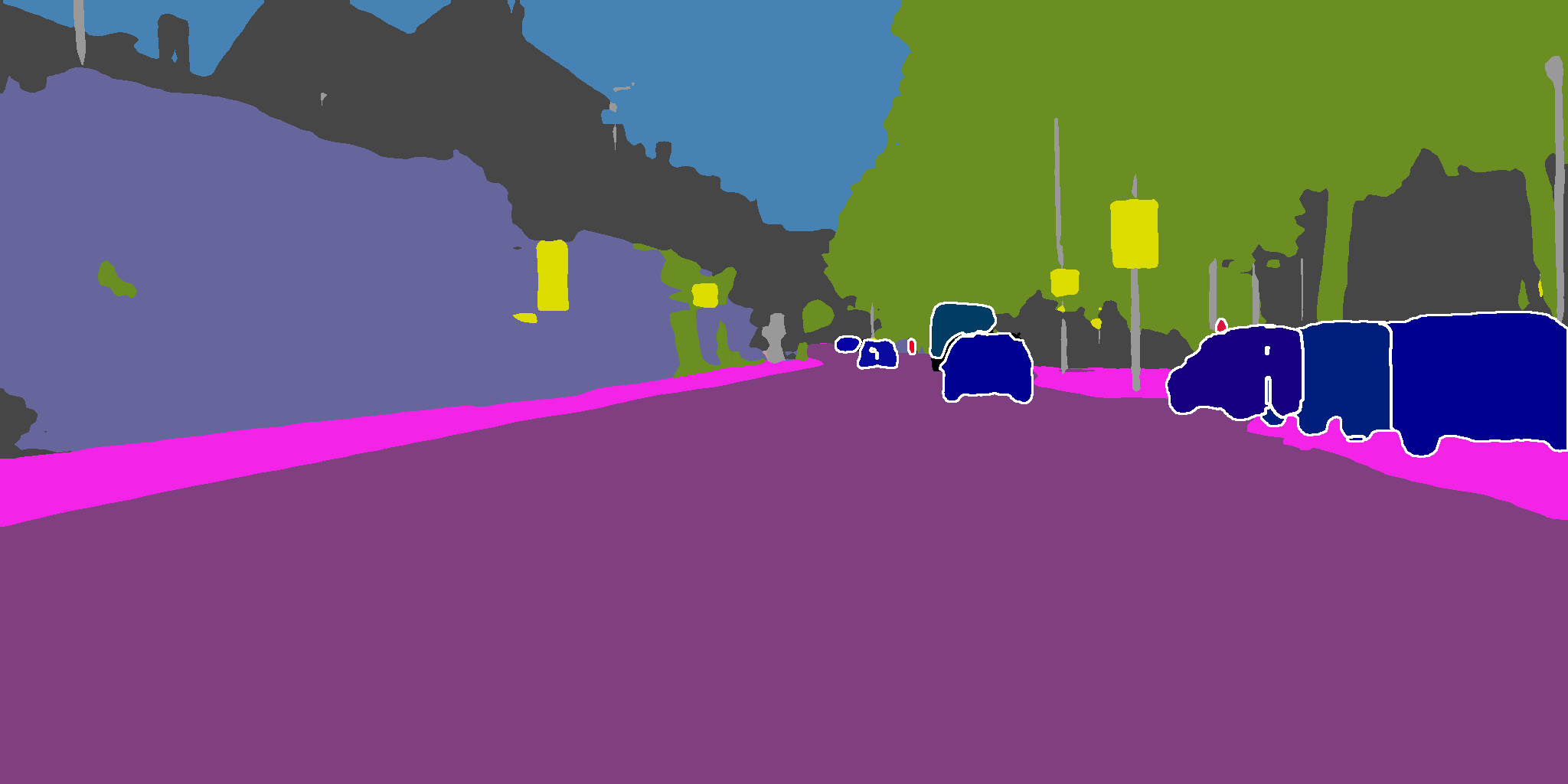}

\includegraphics[width=\figwidth\linewidth]{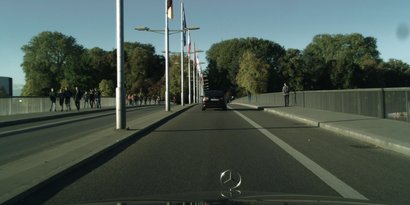}
\includegraphics[width=\figwidth\linewidth]{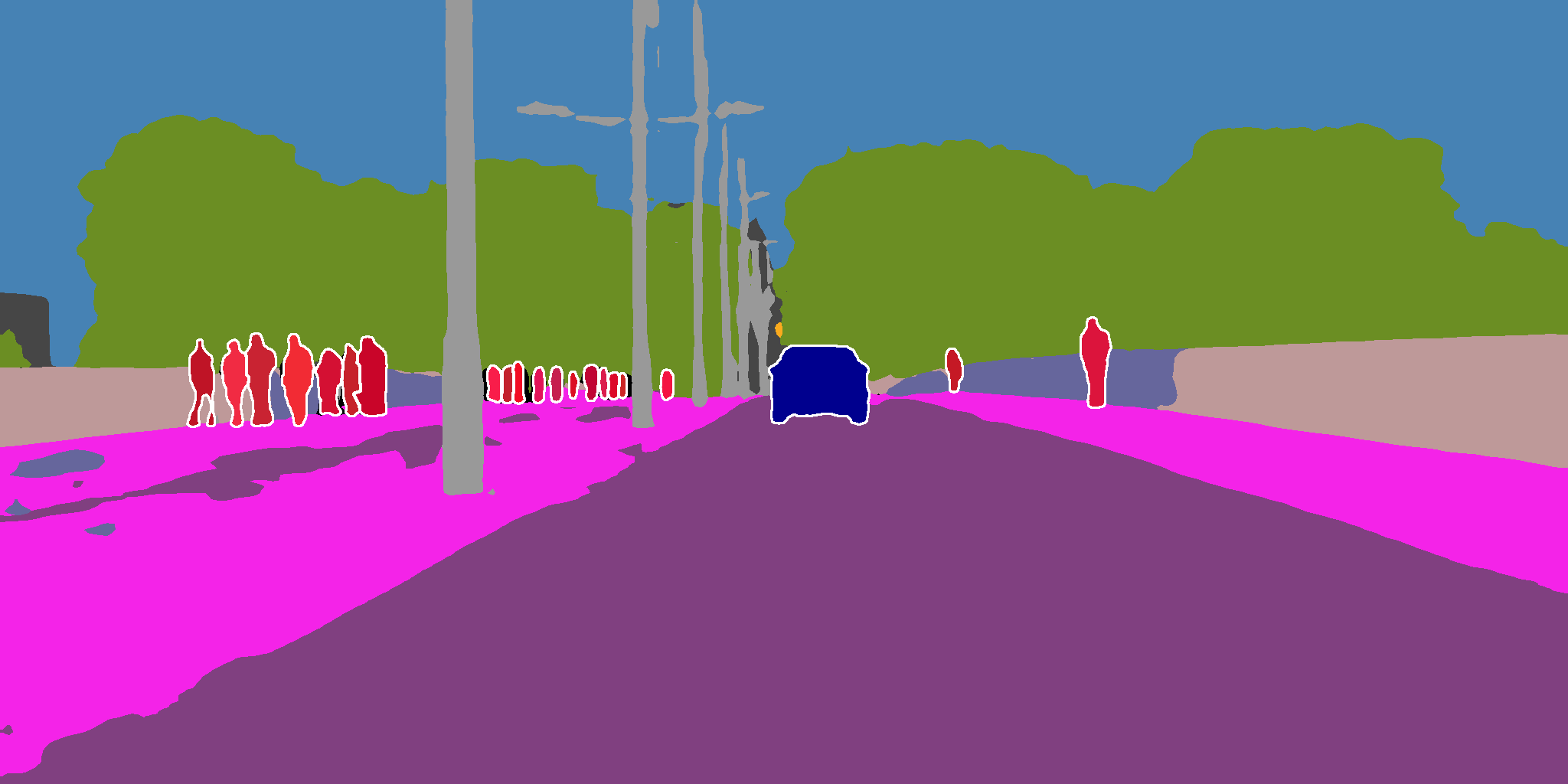}
\end{center}
\vspace{-4mm}
   \caption{\textbf{Weakly supervised panoptic segmentation} Our proposed algorithm obtain promising and practical prediction results, trained with only bounding box and semantic labels.}
\label{fig:weakly-example}
\end{figure}

These labels present a weaker supervision for the panoptic segmentation task, as there are ambiguous pixels between overlapping bounding boxes of the same category.
% However, these labels are easier to obtain as they doesn't require precise separation of the boundary between overlapping instance. 

In our proposed algorithm, we directly regress the bounding box and semantic classification. The construction of instance masks relies on the accuracy of the bounding box predictions instead of the explicit modeling of instance shapes. Thus, our method is robust to the absence of foreground mask information. In our implementation, we relax the pixel assignment during bounding box prediction as depicted in \figurename\ref{fig:weak-label}, favoring the smallest bounding box in overlapping cases.

As shown in Table 3, our weakly supervised model achieved $55.7$ PQ, i.e.~$95\%$ of the performance of the fully supervised model. Some qualitative examples are also provided in \figurename\ref{fig:weakly-example}.

\section{Degradation Curves}
% Please add the following required packages to your document preamble:
% \usepackage{graphicx}
\begin{table}[t]
\centering
\resizebox{\linewidth}{!}{%
\begin{tabular}{c|c|c|c|c|c|c}
\toprule
% Method & 
Backbone & PQ & $\text{PQ}^{th}$ & $\text{PQ}^{st}$ & mIoU & AP & Inference Time \\ \hline
\multicolumn{7}{c}{\textbf{Full Image Size: 1024 $\times$ 2048 px}} \\ \hline
% FPSNet~\cite{de2019fast} & ResNet-50-FPN & 55.1 & 48.3 & 60.1 & - & \multicolumn{1}{c|}{-} & TitanRTX & 114ms \\ \hline
% SSAP~\cite{gao2019ssap} & ResNet-50 & 56.6 & 49.2 & - & - & \multicolumn{1}{c|}{31.5} & 1080Ti & $>$260ms \\ \hline
% DeeperLab~\cite{deeperlab2019} & Light Wider MNV2 & 48.1 & - & - & - & \multicolumn{1}{c|}{-} & V100 & 312ms \\ \hline
% % should be 149 ms
% DeeperLab~\cite{deeperlab2019} & Wider MNV2 & 52.3 & - & - & - & \multicolumn{1}{c|}{-} & V100 & 251ms \\ \hline
% % should be 308 ms
% DeeperLab~\cite{deeperlab2019} & Xception-71 & 56.5 & - & - & - & \multicolumn{1}{c|}{-} & V100 & 312ms \\ \hline
% source:a7n1yk62
ResNet-18-FPN  & 55.5 & 47.6 & 61.2 & 74.7 & 26.7 & 77ms\\ \hline
ResNet-34-FPN  & 56.7 & 48.0 & 62.9 & 75.1 & 26.9 & 83ms\\ \hline
ResNet-50-FPN  & 58.8 & 52.1 & 63.7 & 76.8 & 31.0 & 99ms\\ \hline
\multicolumn{7}{c}{\textbf{Half Image Size: 512 $\times$ 1024 px}} \\ \hline
ResNet-18-FPN  & 46.7 & 37.7 & 53.2 & 69.2 & 18.4 & 40ms\\ \hline
ResNet-34-FPN  & 47.5 & 37.9 & 54.4 & 69.7 & 18.7 & 43ms\\ \hline
ResNet-50-FPN  & 49.5 & 40.7 & 55.9 & 71.0 & 20.9 & 49ms\\ \hline
\hline
\end{tabular}%
}
\caption{\textbf{Performance of Our Proposed Framework with Different Configuration on CityScapes}}
\label{tab:perf_diff_backbone}
\end{table}

In the experiment section of the main text, we provide performance analysis with the best performing configuration of our models.
However, in real-world systems, it is a common practice to apply light-weight backbones and low-resolution input images for better inference efficiency at the expense of accuracy.
In this section, we provide a wider spectrum of the expected performance of our proposed method when working with downgraded configurations. 
We train our proposed framework with multiple ResNet-FPN backbones of different depths. 
More specifically, we replace our default ResNet-50-FPN backbone with ResNet-34-FPN and ResNet-18-FPN, and train these two models using the same set of hyper-parameters (e.g. learning rate) as the default one. 
The two shallower backbones are also pre-trained with the ImageNet dataset. We test the trained models on the CityScapes dataset and report the results in terms of accuracy and inference speed. 
The results are presented in \tablename\ref{tab:perf_diff_backbone}.
As expected, shallower backbones decrease model performance especially for instance segmentation, but further accelerates inference speed.
We also train and evaluate our models on CityScapes with images of half of the original size, namely 512 $\times$ 1024 px. We report the performance in \tablename\ref{tab:perf_diff_backbone} as well.

\begin{table}[ht]
\centering
\resizebox{0.42\textwidth}{!}{%
\begin{tabular}{@{}rccc|rccc@{}}
\toprule
{Class} & {PQ} & {SQ} & {RQ}&{Class} & {PQ} & {SQ} & {RQ} \\ \midrule
\textbf{Mean} & 37.13 & 76.14   &46.98 &tv & 61.49 & 84.71 & 72.59 \\
 bed & 52.43 & 83.47 & 62.82 & bus & 66.56 & 86.39 & 77.05 \\
 car & 45.06 & 76.82 & 58.66 & cat & 71.66 & 86.64 & 82.71 \\
 cow & 53.58 & 77.06 & 69.53 & cup & 43.82 & 81.58 & 53.72 \\
 dog & 62.20 & 83.26 & 74.70 & net & 40.53 & 77.81 & 52.08 \\
 sea & 72.62 & 89.61 & 81.03 & tie & 23.70 & 69.93 & 33.89 \\
 bear & 73.45 & 85.47 & 85.94 & bird & 34.53 & 75.04 & 46.01 \\
 boat & 28.71 & 69.99 & 41.02 & book & 12.43 & 66.94 & 18.57 \\
 bowl & 39.12 & 79.67 & 49.10 & cake & 41.43 & 80.86 & 51.23 \\
 fork & 13.32 & 67.63 & 19.70 & kite & 33.84 & 71.18 & 47.54 \\
 oven & 47.03 & 80.87 & 58.15 & road & 52.09 & 81.22 & 64.14 \\
 roof & 12.68 & 66.09 & 19.19 & sand & 50.00 & 85.77 & 58.29 \\
 sink & 46.48 & 79.31 & 58.60 & skis & 4.17 & 59.49 & 7.02 \\
 snow & 79.12 & 90.81 & 87.12 & tent & 8.00 & 68.01 & 11.76 \\
 vase & 40.49 & 76.72 & 52.77 & apple & 24.08 & 77.40 & 31.11 \\
 bench & 21.75 & 75.45 & 28.83 & chair & 29.38 & 73.59 & 39.92 \\
 clock & 58.55 & 82.25 & 71.19 & couch & 46.16 & 83.70 & 55.16 \\
 donut & 47.41 & 84.45 & 56.13 & fruit & 4.42 & 56.74 & 7.79 \\
 horse & 55.45 & 77.40 & 71.64 & house & 17.33 & 68.41 & 25.33 \\
 knife & 7.36 & 71.53 & 10.28 & light & 15.48 & 67.57 & 22.91 \\
 mouse & 59.03 & 82.46 & 71.58 & pizza & 55.86 & 84.97 & 65.74 \\
 river & 43.97 & 86.54 & 50.81 & sheep & 48.27 & 76.27 & 63.29 \\
 shelf & 13.45 & 62.14 & 21.65 & spoon & 3.95 & 70.62 & 5.59 \\
 towel & 21.27 & 75.28 & 28.25 & train & 69.25 & 87.58 & 79.07 \\
 truck & 37.23 & 79.59 & 46.78 & zebra & 62.10 & 79.77 & 77.85 \\
 banana & 23.25 & 74.59 & 31.17 & banner & 12.18 & 72.80 & 16.72 \\
 bottle & 40.52 & 76.06 & 53.27 & bridge & 12.35 & 62.64 & 19.72 \\
 carrot & 20.33 & 70.37 & 28.89 & flower & 17.92 & 78.22 & 22.92 \\
 gravel & 13.75 & 70.55 & 19.49 & laptop & 52.07 & 80.26 & 64.88 \\
 orange & 29.22 & 81.95 & 35.66 & person & 55.63 & 76.64 & 72.58 \\
 pillow & 1.34 & 67.86 & 1.98 & remote & 22.41 & 74.18 & 30.22 \\
 stairs & 12.77 & 68.65 & 18.60 & toilet & 68.16 & 86.37 & 78.92 \\
 bicycle & 31.45 & 70.83 & 44.40 & blanket & 4.92 & 66.76 & 7.37 \\
 counter & 18.58 & 69.66 & 26.67 & curtain & 42.60 & 78.10 & 54.55 \\
 frisbee & 57.23 & 81.01 & 70.65 & giraffe & 63.30 & 78.75 & 80.38 \\
 handbag & 15.24 & 72.38 & 21.05 & hot dog & 32.36 & 82.80 & 39.08 \\
 toaster & 19.90 & 64.69 & 30.77 & airplane & 60.55 & 80.01 & 75.68 \\
 backpack & 18.91 & 75.04 & 25.20 & broccoli & 29.18 & 72.30 & 40.35 \\
 elephant & 65.76 & 80.56 & 81.62 & keyboard & 49.18 & 80.80 & 60.87 \\
 platform & 19.08 & 79.62 & 23.96 & railroad & 44.68 & 72.61 & 61.54 \\
 sandwich & 33.86 & 81.97 & 41.31 & scissors & 31.51 & 75.92 & 41.51 \\
 suitcase & 42.38 & 78.90 & 53.71 & umbrella & 51.68 & 79.63 & 64.90 \\
 cardboard & 16.68 & 68.63 & 24.31 & microwave & 57.83 & 82.50 & 70.10 \\
 snowboard & 19.16 & 72.23 & 26.53 & stop sign & 67.28 & 91.11 & 73.85 \\
 surfboard & 39.52 & 74.85 & 52.80 & wall-tile & 44.76 & 77.36 & 57.86 \\
 wall-wood & 19.96 & 72.12 & 27.67 & cell phone & 32.06 & 80.52 & 39.81 \\
 door-stuff & 20.98 & 72.71 & 28.85 & floor-wood & 43.09 & 79.43 & 54.25 \\
 hair drier & 0.00 & 0.00 & 0.00 & motorcycle & 47.36 & 76.38 & 62.01 \\
 rug  & 37.32 & 78.32 & 47.65 & skateboard & 44.63 & 71.17 & 62.71\\ 
 teddy bear & 52.60 & 81.06 & 64.90 & toothbrush & 10.08 & 69.57 & 14.49 \\
 wall-brick & 27.88 & 72.79 & 38.31 & wall-stone & 17.45 & 76.97 & 22.67 \\
 wine glass & 35.82 & 76.88 & 46.59 & dirt  & 30.02 & 77.54 & 38.72 \\
 rock  & 32.14 & 77.30 & 41.57 & sports ball & 45.27 & 78.09 & 57.97 \\
 tree  & 64.41 & 80.82 & 79.70 & water-other & 22.09 & 81.91 & 26.97 \\
 baseball bat & 23.83 & 66.73 & 35.71 & dining table & 29.18 & 74.86 & 38.98 \\
 fence  & 26.39 & 71.53 & 36.89 & fire hydrant & 65.68 & 83.51 & 78.65 \\
 grass  & 54.96 & 82.77 & 66.40 & mirror-stuff & 26.98 & 73.91 & 36.50 \\
 paper  & 11.18 & 67.05 & 16.67 & playingfield & 62.58 & 88.66 & 70.59 \\
 potted plant & 28.78 & 71.20 & 40.42 & refrigerator & 57.87 & 85.28 & 67.86 \\
 table  & 20.75 & 72.09 & 28.79 & window-blind & 37.01 & 79.52 & 46.54 \\
 window-other & 27.50 & 72.53 & 37.92 & parking meter & 54.27 & 81.41 & 66.67 \\
 tennis racket & 57.19 & 78.94 & 72.45 & traffic light & 38.18 & 74.12 & 51.51 \\
 baseball glove & 36.60 & 76.41 & 47.90 & cabinet  & 40.15 & 77.20 & 52.01 \\
 ceiling  & 50.09 & 78.23 & 64.03 & mountain  & 41.87 & 75.86 & 55.20 \\
 pavement  & 36.49 & 77.79 & 46.90 & sky-other  & 81.65 & 90.76 & 89.96 \\
 food-other  & 13.83 & 72.99 & 18.95 & wall-other  & 44.06 & 77.33 & 56.97\\ 
 floor-other  & 38.16 & 78.39 & 48.68 & building-other  & 38.17 & 77.90 & 49.00\\ 
\bottomrule
\end{tabular}
}
\caption{\textbf{Per-class Performance on COCO}}
\label{tab:coco-detail}
\end{table}

\clearpage
{\small
\bibliographystyle{ieee_fullname}
\bibliography{egbib}

\begin{thebibliography}{10}
\bibitem{answer}
Ours results are in the middle for row 2,5 and on the right for row 1,3,4.
\end{thebibliography}
}

\end{document}